\documentclass[conference]{IEEEtran}
\usepackage{cite}
\usepackage{amsmath,amssymb,amsfonts}
\usepackage[ruled,linesnumbered]{algorithm2e}
\usepackage{graphicx}
\usepackage{textcomp}
\usepackage[table,xcdraw]{xcolor}
\usepackage{multirow}
\usepackage{xspace}
\usepackage{pifont}
\usepackage{svg}
\usepackage{dsfont}
\usepackage{subcaption}
\usepackage[hidelinks]{hyperref}
\usepackage{cleveref}
\usepackage{booktabs}
\usepackage{balance}
\usepackage{stfloats}
\usepackage{caption}

\def\BibTeX{{\rm B\kern-.05em{\sc i\kern-.025em b}\kern-.08em
    T\kern-.1667em\lower.7ex\hbox{E}\kern-.125emX}}

\newcommand{\no}{\text{\ding{55}}}%
\newcommand{\unknown}{N/A\xspace}%
\newcommand{\yes}{\checkmark}%

\begin{document}

\title{Attackers Can Do Better: Over- and Understated Factors of Model Stealing Attacks
}

\author{
\IEEEauthorblockN{Daryna Oliynyk}
\IEEEauthorblockA{CDL AsTra, \\
Faculty of Computer Science, \\
University of Vienna, Vienna, Austria\\
daryna.oliynyk@univie.ac.at
}
\and
\IEEEauthorblockN{Rudolf Mayer}
\IEEEauthorblockA{\textit{SBA Research}\\
Vienna, Austria\\
rmayer@sba-research.org
}
\and
\IEEEauthorblockN{Andreas Rauber}
\IEEEauthorblockA{Institute of Information Systems Engineering,\\
Faculty of Informatics, \\
TU Wien, Vienna, Austria\\
andreas.rauber@tuwien.ac.at
}
}

\maketitle

\begin{abstract}
Machine learning (ML) models were shown to be vulnerable to model stealing attacks, which lead to intellectual property infringement. Among other attack methods, substitute model training is an all-encompassing attack applicable to any machine learning model whose behaviour can be approximated from input-output queries.
Whereas previous works mainly focused on improving the performance of substitute models by, e.g. developing a new substitute training method, there have been only limited ablation studies that try to understand the impact the strength of an attacker has on the substitute model's performance. As a result, different authors came to diverse, sometimes contradicting, conclusions.
In this work, we exhaustively examine the ambivalent influence of different factors resulting from varying the attacker's capabilities and knowledge on a substitute training attack. 

Our findings suggest that some of the factors that have been considered important in the past are, in fact, not that influential; instead, we discover new correlations between attack conditions and success rate. 
In particular, we demonstrate that better-performing target models enable higher-fidelity attacks and explain the intuition behind this phenomenon. Further, we propose to shift the focus from the complexity of target models toward the complexity of their learning tasks.  Therefore, for the substitute model, rather than aiming for a higher architecture complexity, we suggest focusing on getting data of higher complexity and an appropriate architecture. Finally, we demonstrate that even in the most limited data-free scenario, there is no need to overcompensate weak knowledge with unrealistic capabilities in the form of millions of queries. 
Our results often exceed or match the performance of previous attacks that assume a stronger attacker, suggesting that these stronger attacks are likely endangering a model owner's intellectual property to a significantly higher degree than shown until now. 
\end{abstract}


\section{Introduction}
The rapid evolution of Machine Learning (ML) has added significant value to ML-based solutions, especially to the models they are built upon. However, model owners might need to expose their models to third parties, for instance, through APIs, hence endangering their intellectual property to so-called model stealing (extraction) attacks \cite{tramer_stealing_2016}. 
As one form of this attack, a malicious end user can exploit the interaction channel with the model to collect labelled data and train a \textit{substitute} model to behave similarly to the original \textit{target} model \cite{orekondy_knockoff_2019}. As a result, malicious third parties can possess an illegitimate (approximate) copy of the original model, violating the intellectual property rights of the model owner and endangering their business model.

Substitute model attacks are suitable for targeting models that are vital in numerous fields, including image classification \cite{orekondy_knockoff_2019}, image-to-image translation \cite{szyller_good_2021}, natural language processing \cite{krishna_thieves_2020}, and reinforcement learning \cite{chen_stealing_2021}. 
In addition to being applicable to various task and data domains, substitute training attacks have been demonstrated to be effective under varied assumptions of the attacker's strength. These assumptions primarily relate to the attacker's knowledge of the target model and its training data, as well as the attacker's capabilities, primarily regarded as the number of queries needed to steal a model. 
However, the influence of these factors has been studied only sparsely, with most works addressing a limited subset tailored to their specific attack methods. As a consequence, a certain bias occurred in what factors are considered decisive when an attacker's strength is limited in a certain way. Moreover, as findings from different works sometimes disagree, the impact of some factors remains equivocal.

In this work, we address this gap and systematically examine the influence of different factors on the performance of the substitute model. We demonstrate that while some factors attracted a lot of attention, others have been under-explored or completely overlooked. We present new insights that not only can provide a more comprehensive explanation of observations from previous work but also make previous attacks more severe. 

To this end, we comprehensively evaluate how the performance of a substitute model is influenced by (i) target model properties such as architecture and performance on the original classification task, (ii) substitute architecture choice, (iii) usage of transfer learning for training target and substitute models, (iv) attacker's data quality, and (v) attacker's capabilities represented by query budget and query optimisation strategies. 
Our conclusions are based on an analysis of 180 attack configurations, each being applied against three target models with different properties.

Our main contributions are:
\begin{itemize}
    \item We are the first to demonstrate that the performance of the target model acts as a bottleneck, limiting the fidelity of substitute models. This stems from the ability of substitute models to replicate the target model's behaviour significantly better on \textit{correctly} predicted labels than on labels \textit{incorrectly} predicted by the target model. Consequently, for stronger attackers, achieving high fidelity is easier when attacking better-performing models. 
    \item We propose to shift focus from the target model complexity towards the complexity of its classification task. In particular, the complexity of the substitute model should be considered with respect to the data complexity rather than the complexity of the target model, as suggested in several prior works. 
    \item We introduce novel insights into how the usage of transfer learning impacts the performance of substitute training attacks. In particular, if a substitute model is trained from scratch, it will perform better when stealing a target model trained from scratch than a target model with pre-trained weights. 
    \item We show that attacks are more effective when the complexity of the attacker’s dataset is higher than the original dataset; in contrast, attacks generally perform worse when the attacker's dataset has lower complexity than the original dataset. Moreover, attackers with simpler data tend to overestimate attack performance, whereas attackers with more complex data tend to underestimate it.
    \item Finally, we are the first to perform a data-free attack that is effective with a query budget smaller than the original training set of the target model. 
\end{itemize}

Besides, our substitute models trained on original or problem-domain data outperform state-of-the-art, even when assuming weaker attacker knowledge. Our data-free attack achieves a performance comparable to both data-free attacks and attacks using non-problem domain data while requiring two orders of magnitude fewer queries.

The rest of the paper is organised as follows. \Cref{sec:background} introduces relevant terminology and metrics used in this work for attack evaluation. \Cref{sec:related_work} provides a comprehensive overview of related work on stealing image classifiers. In \Cref{sec:factors}, we present our detailed analysis of factors influencing the attack success. \Cref{sec:comparison_sota} demonstrates a comparison of our results to the state-of-the-art, followed by a discussion in \Cref{sec:discussion} and a conclusion in \Cref{sec:conclusion}.

\section{Background}
\label{sec:background}
The general goal of a model stealing (or model extraction) attack is to create an exact or approximate replica of a machine learning model to which the attacker has some sort of restricted access.
We call the model under attack the \textit{target model} and denote it as $f$. We further assume that we only have \textit{black-box} access to the target model, which means that the only information an adversary can retrieve is the predictions (\textit{outputs}) for given input samples (\textit{inputs}). Moreover, in the context of the classification tasks considered in this work, we assume that the target models can only output top-1 predictions (\textit{labels}) for input samples. In other words, the only available action for the adversary is to send a sample $x$ to the target model and obtain a label $f(x) \in \{c_1, \ldots, c_k\}$. We call such a single request to the model a \textit{query}, where $x$ is the \textit{query input} and $f(x)$ is the \textit{query output}. The adversary can create an \textit{attacker's dataset}, exploiting the target model as an oracle for labelling data samples. 
Subsequently, the attacker's dataset can be used to train a so-called \textit{substitute model}. The substitute model, which we denote as $\hat{f}$, can then further be used to launch an alternative service with lower fees, potentially leading to lower demand for the original service and profit loss for the target model owner. 

A successfully performed attack should be both efficient and effective. In terms of efficiency, an attacker should spend a reasonable amount of resources to collect and label the data through the target model. The most crucial part of this process is the \textit{number of queries} required to perform the attack. 
Previous work suggested reporting an \textit{efficiency score}, which shows how many queries per parameter (weight) of the target model it takes to perform an attack \cite{oliynyk_i_2023}. In this work, we shift focus from the target model's complexity towards the complexity of its training data and, therefore, measure how many queries per target's training sample are required to perform the attack. 

For evaluating effectiveness, the performance of a substitute model is compared with the performance of the target model. Three metrics are mainly used for that purpose: accuracy, fidelity, and transferability \cite{oliynyk_i_2023}. In the following, we briefly describe the metrics used in this work.

\textit{Accuracy} shows how a substitute model performs on the classification task that the target model was trained to solve. It compares the outputs of the substitute model with the original labels of a dataset. The accuracy of the substitute model $\hat{f}$ on dataset $X^{test}$ is measured as 
$$\dfrac{1}{|X^{test}|} \sum_{i=1}^{|X^{test}|} \mathds{1}_{(\hat{f}(x_i^{test}) = y_{i}^{test})}.$$
Additionally, we introduce in this work a metric called \textit{joint accuracy}, which shows how many samples are classified correctly by both target $f$ and substitute $\hat{f}$ models, as follows
$$\dfrac{1}{|X^{test}|} \sum_{i=1}^{|X^{test}|} \mathds{1}_{(f(x_i^{test}) = \hat{f}(x_i^{test}) = y_{i}^{test})}.$$
\textit{Fidelity} measures the similarity of target and substitute predictions by comparing the labels that the target model $f$ and the substitute model $\hat{f}$ output on the test set $X^{test}$. Therefore, fidelity counts both correct and incorrect equal predictions, as
$$\dfrac{1}{|X^{test}|} \sum_{i=1}^{|X^{test}|} \mathds{1}_{(\hat{f}(x_i^{test}) = f(x_i^{test}))}.$$

\section{Related work}
\label{sec:related_work}
\begin{table*}[ht!]
\centering
\small
\caption{Summary of current substitute training approaches.}
\begin{tabular}{lllccll}
\toprule
\textbf{Paper}                        & \textbf{Attacker's data}   & \textbf{Data crafting technique} & \textbf{\begin{tabular}[c]{@{}c@{}}Target\\ architecture\end{tabular}} & \textbf{\begin{tabular}[c]{@{}c@{}}Query\\ optimization\end{tabular}} & \textbf{Target outputs} & \textbf{Metrics} \\ \midrule
\cite{orekondy_knockoff_2019}         & NPD                        &                                  & \yes \no                                                               & RL                                                                    & Labels, Probab.         & Acc              \\
\rowcolor[HTML]{EFEFEF} 
\cite{jagielski_high_2020}            & Original or NPD            &                                  & \no                                                                    & SSL                                                                   & Probab.                 & Acc, Fid         \\
\cite{papernot_practical_2017}        & Original or PD             & Adver. augm.                     & \no                                                                    &                                                                       & Labels                  & Acc              \\
\rowcolor[HTML]{EFEFEF} 
\cite{correia-silva_copycat_2018}     & NPD or/and PD              &                                  & \unknown                                                               &                                                                       & Labels                  & Acc              \\
\cite{pal_activethief_2020}           & NPD                        &                                  & \yes \no                                                               & AL                                                                    & Labels, Probab.         & Fid              \\
\rowcolor[HTML]{EFEFEF} 
\cite{juuti_prada_2019}               & Original                   & Adver. augm.                     & \yes \no                                                               &                                                                       & Labels, Probab.         & Fid              \\
\cite{atli_extraction_2020}           & NPD                        &                                  & \yes                                                                   & RL                                                                    & Labels, Probab.         & Acc              \\
\rowcolor[HTML]{EFEFEF} 
\cite{pengcheng_query-efficient_2018} & Original                   & Adver. augm.                     & \no                                                                    & AL                                                                    & \unknown                & Acc, Fid         \\
\cite{mosafi_stealing_2019}           & NPD                        & Data composition                 & \no                                                                    &                                                                       & Labels                  & Acc              \\
\rowcolor[HTML]{EFEFEF} 
\cite{yuan_es_2022}                   & Artificial                 & Generator                        & \yes \no                                                               &                                                                       & Labels, Probab.         & Acc              \\
\cite{milli_model_2019}               & Original                   &                                  & \yes \no                                                               &                                                                       & Gradients               & Acc              \\
\rowcolor[HTML]{EFEFEF} 
\cite{kariyappa_maze_2021}            & Artificial                 & Generator                        & \no                                                                    &                                                                       & Probab.                 & Acc              \\
\cite{roberts_model_2019}             & Artificial                 & Noise                            & \yes                                                                   &                                                                       & Probab.                 & Acc              \\
\rowcolor[HTML]{EFEFEF} 
\cite{barbalau_black-box_2020}        & NPD                        & Generator                        & \yes \no                                                               & EA                                                                    & Probab.                 & Acc              \\
\cite{yu_cloudleak_2020}              & PD                         & Adver. augm.                     & \yes \no                                                               &                                                                       & Probab.                 & Acc              \\
\rowcolor[HTML]{EFEFEF} 
\cite{gong_inversenet_2021}           & NPD                        & Model inversion                  & \no                                                                    &                                                                       & Labels                  & Acc, Fid         \\
\cite{truong_data-free_2021}          & Artificial                 & Generator                        & \no                                                                    &                                                                       & Probab.                 & Acc              \\
\rowcolor[HTML]{EFEFEF} 
\cite{miura_megex_2024}               & Artificial                 & Generator                        & \no                                                                    &                                                                       & Probab. + Expl.         & Acc              \\
\cite{sanyal_towards_2022}            & Artificial                 & Generator                        & \no                                                                    &                                                                       & Labels                  & Acc              \\
\rowcolor[HTML]{EFEFEF} 
\cite{zhang_thief_2021}               & Original                   & Adver. augm.                     & \yes \no                                                               & RL                                                                    & Probab.                 & Acc              \\
\cite{wang_enhance_2022}              & NPD                        &                                  & \unknown                                                               & AL                                                                    & Labels                  & Acc              \\
\rowcolor[HTML]{EFEFEF} 
\cite{wang_black-box_2022}            & NPD                        &                                  & \yes \no                                                               &                                                                       & Labels                  & Acc, Fid         \\
\cite{yan_towards_2022}               & Original                   &                                  & \no                                                                    &                                                                       & Probab. + Expl.         & Acc              \\
\rowcolor[HTML]{EFEFEF} 
\cite{xie_game_2022}                  & (N)PD                      & Generator                        & \no                                                                    & AL                                                                    & Probab.                 & Acc, Fid         \\
\cite{chen_d-dae_2023}                & Original, PD or NPD        &                                  & \yes                                                                   & AL, RL                                                                & Probab.                 & Acc, Fid         \\
\rowcolor[HTML]{EFEFEF} 
\cite{he_drmi_2021}                   & Original or NPD            &                                  & \yes                                                                   & DR                                                                    & Probab.                 & Acc              \\
\cite{liu_ml-doctor_2022}             & Original                   &                                  & \yes                                                                   &                                                                       & Probab.                 & Acc, Fid         \\
\rowcolor[HTML]{EFEFEF} 
\cite{rosenthal_disguide_2023}        & Artificial                 & Generator                        & \yes \no                                                               &                                                                       & Labels, Probab.         & Acc              \\
\cite{yan_holistic_2023}              & Original, PD or NPD        &                                  & \yes \no                                                               &                                                                       & Labels                  & Acc, Fid         \\
\rowcolor[HTML]{EFEFEF} 
\cite{zhang_towards_2022}             & Artificial                 & Generator                        & \no                                                                    &                                                                       & Labels, Probab.         & Acc              \\
\cite{yan_explanation-based_2023}     & Artificial                 & Generator                        & \no                                                                    &                                                                       & Labels + Expl.          & Acc              \\
\rowcolor[HTML]{EFEFEF} 
\cite{yan_explanation_2023}           & Original                   &                                  & \unknown                                                               &                                                                       & Labels + Expl.          & Fid              \\
\cite{yang_efficient_2023}            &Artificial         & Generator                        & \no                                                                    &                                                                       & Labels                  & Acc              \\
\rowcolor[HTML]{EFEFEF} 
\cite{lin_quda_2023}                  & Artificial         & Adver. augm., Generator          & \yes \no                                                               & RL                                                                    & Labels, Probab.         & Acc              \\
\cite{liu_shrewdattack_2023}          & Original and PD            &                                  & \unknown                                                               & CBS                                                                   & Probab.                 & Acc              \\
\rowcolor[HTML]{EFEFEF} 
\cite{pape_limitations_2023}          & Original                   &                                  & \yes \no                                                               &                                                                       & Labels                  & Fid              \\
\cite{liu_efficient_2024}            & Artificial                 & Generator                        & \no                                                                    &                                                                       & Probab.                 & Acc, Fid         \\
\rowcolor[HTML]{EFEFEF} 
\cite{khaled_careful_2022}            & NPD                        &                                  & \no                                                                    & RL                                                                    & Probab.                 & Acc, Fid         \\
\cite{zhao_extracting_2023}           & Original, PD or NPD        &                                  & \no                                                                    & SSL                                                                   & Probab.                 & Fid              \\
\rowcolor[HTML]{EFEFEF} 
\cite{karmakar_marich_2023}           & NPD                        &                                  & \no                                                                    & AL                                                                    & Labels                  & Acc, Fid         \\
\cite{jindal_army_2024}        & NPD                        &                                  & \no                                                                    & AL, SSL                                                               & Labels                  & Acc, Fid         \\
\rowcolor[HTML]{EFEFEF} 
\cite{beetham_dual_2023}              & Artificial                 & Generator                        & \no                                                                    &                                                                       & Labels, Probab.         & Acc              \\
\cite{hong_exploring_2023}            & Artificial                 & Generator                        & \no                                                                    &                                                                       & Labels, Probab.         & Acc              \\
\rowcolor[HTML]{EFEFEF} 
\cite{yang_swifttheft_2024}         & NPD                        &                                  & \yes                                                                   & AL                                                                    & Labels                  & Fid              \\
\textbf{Our work}                     & Original, PD or Artificial & Adver. augm., Generator          & \yes \no                                                               & AL                                                                    & Labels                  & Acc, Fid        \\ \bottomrule
\end{tabular}
\label{tab:substitute_attacks_comparison}
\end{table*}

In this section, we discuss aspects and approaches relevant to the area of stealing image classifiers. \Cref{tab:substitute_attacks_comparison} shows aggregated information from 44 relevant papers, which were selected based on the following criteria: (i) the paper introduces a new substitute training attack or extends a previous work; (ii) both target and substitute models are trained on image data; (iii) the substitute model is trained with the intention of copying the behaviour (functionality) of the target model. In the following, we review each characteristic presented in \Cref{tab:substitute_attacks_comparison}.

\textbf{Attacker's data} corresponds to data categories used for querying the target model; they comprise original, problem-domain, non-problem domain, and artificial data \cite{oliynyk_i_2023}. Attacks that only rely on artificial data to train substitute models are also called \textit{data-free} attacks.  In \Cref{tab:substitute_attacks_comparison}, we indicate in the corresponding column which data was used by related work. We use the connector "and" to mark cases when a substitute model was trained on a combination of data from different categories. 
If a paper assumed that a small amount of original or PD data is available (even if less than 5\%), we still considered those attacks as ones requiring the corresponding type of data.  

\textbf{Data crafting techniques} are commonly used for two goals: generating artificial data and creating more (high-quality) data from the one available. Below, we describe the methods from previous work, which are listed in \Cref{tab:substitute_attacks_comparison}. 
\textit{Adversarial augmentation} is the most common approach for improving the quality of the data \cite{pengcheng_query-efficient_2018, yu_cloudleak_2020, zhang_thief_2021}. The idea is to query a target model with adversarial examples crafted for a substitute model to "correct" the predictions of the substitute model near its decision boundary. 
\textit{Data composition} is an approach proposed for increasing the quality of NPD data by merging two images into one \cite{mosafi_stealing_2019}. 
\textit{Model inversion} is inspired by the model inversion attack \cite{fredrikson_model_2015}, which aims to reconstruct training data from a model. In the context of model stealing, it can be applied to extract data from the target model, leading to more meaningful data when only NPD data is available \cite{gong_inversenet_2021}. 
\textit{Generative models} (generators) are common for creating artificial data \cite{truong_data-free_2021} and improving the quality or increasing the quantity of the attacker's data samples \cite{barbalau_black-box_2020}.
Finally, \textit{noise} can also be used as a direct input for querying a model when no data or generative model is available \cite{roberts_model_2019}.

If the \textbf{target architecture} is known, an adversary can use it as the architecture choice for a substitute model, simplifying the whole stealing process. In general, there are two possible scenarios: (i) the substitute architecture differs from the target, as the latest is assumed to be unknown (marked as $\no$ in \Cref{tab:substitute_attacks_comparison}), (ii) the substitute and target architectures are the same (marked as $\yes$ in \Cref{tab:substitute_attacks_comparison}).
We indicate papers with $\yes \no$ if both cases are reported. 
If it was unclear which strategy the authors chose, we marked such papers as \unknown.

\textbf{Query optimisation} includes techniques that aim to increase the efficiency of an attack by reducing the number of queries. The most common technique is using \textit{Active Learning} (AL) \cite{pal_activethief_2020, pengcheng_query-efficient_2018, wang_enhance_2022, xie_game_2022}. 
Active learning was initially introduced as an optimisation for labelling data in supervised learning scenarios with a significant amount of unlabelled data.
Since labelling data is also part of model stealing (with the API being the oracle), active learning can be applied for query optimisation \cite{chandrasekaran_exploring_2020}. A few other works trained \textit{Reinforcement Learning} (RL) agents to pick samples with the highest impact on substitute model training \cite{orekondy_knockoff_2019, zhang_thief_2021}. Besides, query optimisation can also be conducted by utilising \textit{Evolutionary Algorithms} (EAs) \cite{barbalau_black-box_2020}, \textit{Self-Supervised Learning} (SSL) \cite{jagielski_high_2020, zhao_extracting_2023, jindal_army_2024}, \textit{Dataset Reduction} (DR) \cite{chen_d-dae_2023} or \textit{Cluster-Based Selection} (CBD) of data samples \cite{liu_shrewdattack_2023}.

\textbf{Target outputs} are the primary source of information an adversary can obtain from the target model. The most widespread assumptions are that the target model outputs either \textit{labels} or \textit{probabilities} (confidence scores). All papers that compared labels with probabilities concluded that the substitute performance is better when probabilities are used \cite{pal_activethief_2020, juuti_prada_2019, orekondy_knockoff_2019, atli_extraction_2020, yuan_es_2022, gong_inversenet_2021}. In other scenarios, an API can reveal even more information though explicit model \textit{gradients} \cite{milli_model_2019} or model explanations, provided additionally to labels \cite{yan_explanation-based_2023, yan_explanation_2023} or probabilities \cite{miura_megex_2024, yan_towards_2022}. We marked in \Cref{tab:substitute_attacks_comparison} papers as \unknown in case it was unclear from the attack description which target outputs were utilised. 

Finally, for each paper, we specified the \textbf{metrics} used for attack evaluation, namely, \textit{accuracy} or \textit{fidelity}. 

\Cref{tab:substitute_attacks_comparison} illustrates the following notable trends:
\begin{itemize}
    \item PD data is the least explored in previous studies (18\% of works), in contrast to artificial (27\%), original (33\%), and NPD (49\%) data. However, as image data is nowadays widely accessible, gathering PD data for image classification tasks is a highly plausible scenario, making PD attacks potentially the most threatening. As we demonstrate later in \Cref{sec:factor:data}, PD data, even of noticeably higher complexity than the original training data, leads to highly effective model stealing attacks. 
    \item Most of the methods were validated with mismatching substitute and target architectures, suggesting that knowledge about the target model architecture is not crucial. This observation aligns with our insight from \Cref{sec:factor:substitute_architecture} that the substitute architecture choice should correspond to the complexity of the attacker's data rather than be compared with the complexity of the target model. 
    \item None of the works that use artificial data incorporate a query optimisation strategy. However, as we demonstrate later in \Cref{sec:comparison_sota}, these attacks usually require millions of queries, which makes them significantly less efficient than attacks using non-artificial data. 
\end{itemize}

\section{Attack Factor Evaluation} \label{sec:factors}
In the following, we first present an overview of all attack configurations studied in this work. The overview is followed by a step-by-step analysis of five attack factors. For each factor, we first present relevant findings from prior work together with our observations, which were not necessarily highlighted or spotted by corresponding authors. Subsequently, we describe and analyse experiments conducted in this work that demonstrate the influence of the attack factor. We conclude the analysis of each attack factor with a summary of our insights.  

\subsection{Attack Setup}
\label{sec:attack_setup}
For our study, we first trained three target models and then exploited each of them to train 180 substitute models (see \Cref{appendix:attack_setup} for a detailed attack setup overview). 
The substitute models varied in terms of architecture, the data they were trained on, the amount of data, and the strategy of data collection.
Below, we describe in detail the characteristics of each component of the attack setup.

\textbf{Target model.} The three target models are trained on the CIFAR-10 dataset \cite{krizhevsky_learning_2009}, with their accuracy scores and complexities shown in \Cref{tab:target_models}. Two target models have ResNet-34 architecture: one is trained from scratch, and the other is trained using transfer learning from the ImageNet dataset. The third model has the SimpleNet \cite{hasanpour_lets_2016} architecture, which has significantly fewer parameters (around one-quarter of ResNet-34) and is trained without transfer learning.

\begin{table}[t!]
\centering
\caption{Accuracy of target models trained on CIFAR-10.}
\begin{tabular}{lcr}
\toprule
\textbf{Target model}                         & \textbf{Test accuracy} & \textbf{\#parameters} \\ \midrule
SimpleNet                     & 91.76\%       & $\sim$5M             \\ 
ResNet-34 (from scratch)       & 93.61\%       & $\sim$21M            \\ 
ResNet-34 (transfer learning)  & 97.14\%       & $\sim$21M            \\ 
\bottomrule
\end{tabular}

\label{tab:target_models}
\end{table}

\textbf{Substitute model.} The substitute models have SimpleNet, ResNet-18 and ResNet-34 architectures. The Simplenet model is trained from scratch, whereas both residual networks are trained using transfer learning from ImageNet.

\textbf{Attacker's Dataset.} Three datasets are used as attacker's data, each corresponding to a different degree of knowledge about the original data, namely: CIFAR-10 (as original data), CINIC-10 (as problem-domain data), and an artificial dataset (for a data-free attack scenario). The artificial data is generated with a stable diffusion model\footnote{\url{https://huggingface.co/stabilityai/stable-diffusion-2-1}} to approximate problem domain data (see \Cref{appendix:artifical_dataset_generation} for more details). Each dataset is split into 45,000 training samples used for training substitute models and 5,000 validation samples. In all experiments, we assume that an attacker only has access to the output labels of the target model. 

\textbf{Query Budget.} We carry out experiments with 5 query budgets: 1,000, 5,000, 10,000, 20,000, and 45,000 samples. We set the upper bound to be 45,000 for two reasons: (i) to avoid knowledge leakage, we keep validation sets unseen by the substitute models, and (ii) for attacks using data other than the original, we want to have a baseline trained on the original data with the same number of queries. 

\textbf{Attack Strategy.} Alongside randomly selecting a subset of the inputs from the whole dataset for querying the target model (random sample selection), we applied three query optimisation strategies: active learning from \cite{pal_activethief_2020}, adversarial augmentation from \cite{pengcheng_query-efficient_2018}, and their combination. These methods from previous work were slightly modified for better efficiency and additionally combined into a single method (see \Cref{appendix:query_optimisation_methods} for more details).

\textbf{Evaluation.} We evaluated each attack in terms of accuracy, joint accuracy (see \Cref{sec:background}), and fidelity. The 10,000 test samples from the CIFAR-10 dataset were used for the final evaluation. All attackers' decisions about the optimal hyperparameter choice were based on scores obtained on 5,000 samples from the attacker's validation set.   

\subsection{Factor 1: Target Model Properties}
\label{sec:factor:target_model}
\textbf{Prior Work Discussion.} Jagielski et al. \cite{jagielski_high_2020} demonstrated that high fidelity of substitute models can not be guaranteed due to the non-deterministic nature of learning-based approaches. We take this exploration further, showing that fidelity is actually limited by the accuracy of the target model, and the non-determinism occurs primarily while learning the mistakes (wrongly classified samples) of the target model. This insight can also be traced in other works, even though it remained unnoticed by the authors. For instance, Pape et al. \cite{pape_limitations_2023} performed a substitute training attack with original data against three target models of different complexity and different performance. In their results, fidelity scores correlate with the performance of the target model, being the highest for the most complex and better-performing model. While one might connect such a trend with the complexity of the target model, we demonstrate that among two target models that have identical architectures but different accuracy scores, the better-performing one leads to higher fidelity.

\textbf{Results.} We present in \Cref{tab:attacks_baseline} the results of attacks with the strongest attacker's knowledge assumption: original data and identical architectures for target and substitute models. The fidelity scores correlate with the performance of the target model---the higher the accuracy of the target model, the higher the fidelity score reached by a substitute model. Moreover, fidelity does not exceed the target accuracy, suggesting that the target accuracy acts as a limiting factor. 
\begin{table}[th]
\centering
\caption{Performance of substitute models trained on the original data (CIFAR-10) with the same architectures as target models.}
\resizebox{\columnwidth}{!}{
\begin{tabular}{llllll}
\toprule
                                                                                                             &                                                                          & \multicolumn{3}{c}{\textbf{Test Scores}}                                                                                                                 &                                                                             \\ \cmidrule{3-5}
\multirow{-2}{*}{\textbf{Target model}}                                     & \multirow{-2}{*}{\begin{tabular}[c]{@{}c@{}}\textbf{Query}\\ \textbf{budget}\end{tabular}} & \multicolumn{1}{c}{\textit{Joint Acc}}                                & \multicolumn{1}{c}{\textit{Accuracy}}                                 & \textit{Fidelity}         & \multirow{-2}{*}{\begin{tabular}[c]{@{}c@{}}\textbf{Target}\\ \textbf{accuracy}\end{tabular}} \\ \midrule
                                                                                                             & 1k                                                                       & \multicolumn{1}{c}{48.07\%}                                  & \multicolumn{1}{c}{50.26\%}                                  & 50.39\%          &                                                                             \\ 
                                                                                                             & 5k                                                                       & \multicolumn{1}{c}{69.84\%}                                  & \multicolumn{1}{c}{72.01\%}                                  & 73.01\%          &                                                                             \\ 
                                                                                                             & 10k                                                                      & \multicolumn{1}{c}{76.39\%}                                  & \multicolumn{1}{c}{78.84\%}                                  & 79.93\%          &                                                                             \\ 
                                                                                                             & 20k                                                                      & \multicolumn{1}{c}{82.70\%}                                   & \multicolumn{1}{c}{85.38\%}                                  & 86.33\%          &                                                                             \\ 
\multirow{-5}{*}{SimpleNet}                                                                                  & 45k                                                                      & \multicolumn{1}{c}{\textbf{87.02\%}}                         & \multicolumn{1}{c}{\textbf{90.27\%}}                         & \textbf{90.61\%} & \multirow{-5}{*}{91.76\%}                                                   \\ \addlinespace
\rowcolor[HTML]{FFFFFF} 
\cellcolor[HTML]{FFFFFF}                                                                                     & 1k                                                                       & \multicolumn{1}{c}{\cellcolor[HTML]{FFFFFF}76.69\%}          & \multicolumn{1}{c}{\cellcolor[HTML]{FFFFFF}79.35\%}          & 78.74\%          & \cellcolor[HTML]{FFFFFF}                                                    \\ 
\rowcolor[HTML]{FFFFFF} 
\cellcolor[HTML]{FFFFFF}                                                                                     & 5k                                                                       & \multicolumn{1}{c}{\cellcolor[HTML]{FFFFFF}85.32\%}          & \multicolumn{1}{c}{\cellcolor[HTML]{FFFFFF}88.54\%}          & 87.24\%          & \cellcolor[HTML]{FFFFFF}                                                    \\ 
\rowcolor[HTML]{FFFFFF} 
\cellcolor[HTML]{FFFFFF}                                                                                     & 10k                                                                      & \multicolumn{1}{c}{\cellcolor[HTML]{FFFFFF}88.51\%}          & \multicolumn{1}{c}{\cellcolor[HTML]{FFFFFF}91.91\%}          & 90.51\%          & \cellcolor[HTML]{FFFFFF}                                                    \\ 
\rowcolor[HTML]{FFFFFF} 
\cellcolor[HTML]{FFFFFF}                                                                                     & 20k                                                                      & \multicolumn{1}{c}{\cellcolor[HTML]{FFFFFF}90.73\%}          & \multicolumn{1}{c}{\cellcolor[HTML]{FFFFFF}94.73\%}          & 92.39\%          & \cellcolor[HTML]{FFFFFF}                                                    \\ 
\rowcolor[HTML]{FFFFFF} 
\multirow{-5}{*}{\cellcolor[HTML]{FFFFFF}\begin{tabular}[c]{@{}c@{}}ResNet-34\\ (from scratch)\end{tabular}} & 45k                                                                      & \multicolumn{1}{c}{\cellcolor[HTML]{FFFFFF}\textbf{92.15\%}} & \multicolumn{1}{c}{\cellcolor[HTML]{FFFFFF}\textbf{96.67\%}} & \textbf{93.50\%}  & \multirow{-5}{*}{\cellcolor[HTML]{FFFFFF}93.61\%}                           \\ \addlinespace
                                                                                                             & 1k                                                                       & \multicolumn{1}{c}{80.58\%}                                  & \multicolumn{1}{c}{81.61\%}                                  & 81.79\%          &                                                                             \\ 
                                                                                                             & 5k                                                                       & \multicolumn{1}{c}{89.47\%}                                  & \multicolumn{1}{c}{90.29\%}                                  & 90.91\%          &                                                                             \\ 
                                                                                                             & 10k                                                                      & \multicolumn{1}{c}{91.30\%}                                   & \multicolumn{1}{c}{92.26\%}                                  & 92.73\%          &                                                                             \\ 
                                                                                                             & 20k                                                                      & \multicolumn{1}{c}{93.77\%}                                  & \multicolumn{1}{c}{94.79\%}                                  & 95.16\%          &                                                                             \\ 
\multirow{-5}{*}{\begin{tabular}[c]{@{}c@{}}ResNet-34\\ (transfer learning)\end{tabular}}                    & 45k                                                                      & \multicolumn{1}{c}{\textbf{95.37\%}}                         & \multicolumn{1}{c}{\textbf{96.49\%}}                         & \textbf{96.84\%} & \multirow{-5}{*}{97.14\%}                                                   \\ \bottomrule
\end{tabular}
}
\label{tab:attacks_baseline}
\end{table}

\begin{table}[th]
\centering
\caption{Accuracy of substitute models on correct and incorrect predictions of the target model. Substitutes are trained with 45,000 original samples and have the same architectures as the target models.}
\resizebox{\columnwidth}{!}{\begin{tabular}{lll}
\toprule
\textbf{Target model}                  & \textbf{Correct predictions } & \textbf{Incorrect predictions } \\ \midrule
SimpleNet                     & 94.83\%                     & 43.57\%                        \\ 
ResNet-34 \scriptsize{(from scratch)}       & 98.44\%                     & 21.13\%                        \\ 
ResNet-34 \scriptsize{(transfer learning)} & 98.18\%                     & 51.40\%                        \\ \bottomrule
\end{tabular}
}
\label{tab:correct_incorrect}
\end{table}
To investigate this behaviour further, we measured how well the substitute model learned correct and incorrect predictions of the target model. Identical correct predictions are represented by joint accuracy. We measure the accuracy of the substitute model on correct target predictions as 
$$\dfrac{\text{Joint accuracy}}{\text{Target accuracy}} \times 100\%.$$
Subsequently, for the incorrect predictions, the accuracy is 
$$\dfrac{\text{Fidelity} - \text{Joint accuracy}}{100\% - \text{Target accuracy}} \times 100\%.$$

\begin{table*}[t]
\centering
\caption{Performance of substitute models with different architectures trained on the original (CIFAR-10) data.}
\begin{tabular}{ccccccccccc}
\toprule
                            & \textbf{Target $\rightarrow$}                                                           & \multicolumn{3}{c}{\textbf{SimpleNet}}                                                                                                              & \multicolumn{3}{c}{\begin{tabular}[c]{@{}c@{}}\textbf{ResNet-34} \\ \textbf{(from scratch)}\end{tabular}}                                                  & \multicolumn{3}{c}{\begin{tabular}[c]{@{}c@{}}\textbf{ResNet-34} \\ \textbf{(transfer learning)}\end{tabular}}                                              \\ \cmidrule(lr){3-5} \cmidrule(lr){6-8} \cmidrule(lr){9-11}
\textbf{Substitute $\downarrow$}     &  \textit{QB}  & \textit{Joint Acc} & \textit{Acc}                                      & \textit{Fid}                                      & \textit{Joint Acc} & \textit{Acc}                                      & \textit{Fid}                                     & \textit{Joint Acc} & \textit{Acc}                                      & \textit{Fid}                                      \\ \midrule
                            & 1k                                                                             & \cellcolor[HTML]{EFEFEF}48.07\%                      & \cellcolor[HTML]{EFEFEF}50.26\%          & \cellcolor[HTML]{EFEFEF}50.39\%          & 48.95\%                                             & 50.41\%                                  & 51.01\%                                 & \cellcolor[HTML]{EFEFEF}47.32\%                     & \cellcolor[HTML]{EFEFEF}48.08\%          & \cellcolor[HTML]{EFEFEF}48.16\%          \\
                            & 5k                                                                             & \cellcolor[HTML]{EFEFEF}69.84\%                      & \cellcolor[HTML]{EFEFEF}72.01\%          & \cellcolor[HTML]{EFEFEF}73.01\%          & 70.82\%                                             & 72.28\%                                  & 73.64\%                                 & \cellcolor[HTML]{EFEFEF}70.50\%                      & \cellcolor[HTML]{EFEFEF}71.44\%          & \cellcolor[HTML]{EFEFEF}71.49\%          \\
                            & 10k                                                                            & \cellcolor[HTML]{EFEFEF}76.39\%                      & \cellcolor[HTML]{EFEFEF}78.84\%          & \cellcolor[HTML]{EFEFEF}79.93\%          & 77.88\%                                             & 79.55\%                                  & 81.05\%                                 & \cellcolor[HTML]{EFEFEF}78.14\%                     & \cellcolor[HTML]{EFEFEF}79.13\%          & \cellcolor[HTML]{EFEFEF}79.27\%          \\
                            & 20k                                                                            & \cellcolor[HTML]{EFEFEF}82.70\%                       & \cellcolor[HTML]{EFEFEF}85.38\%          & \cellcolor[HTML]{EFEFEF}86.33\%          & 83.35\%                                             & 85.15\%                                  & 86.55\%                                 & \cellcolor[HTML]{EFEFEF}84.32\%                     & \cellcolor[HTML]{EFEFEF}85.26\%          & \cellcolor[HTML]{EFEFEF}85.55\%          \\
\multirow{-5}{*}{SimpleNet} & 45k                                                                            & \cellcolor[HTML]{EFEFEF}\textbf{87.02\%}             & \cellcolor[HTML]{EFEFEF}\textbf{90.27\%} & \cellcolor[HTML]{EFEFEF}\textbf{90.61\%} & \textbf{88.34\%}                                    & \textbf{90.45\%}                         & \textbf{91.41\%}                        & \cellcolor[HTML]{EFEFEF}\textbf{89.08\%}            & \cellcolor[HTML]{EFEFEF}\textbf{90.12\%} & \cellcolor[HTML]{EFEFEF}\textbf{90.33\%} \\ \addlinespace
                            & 1k                                                                             & 78.54\%                                              & 82.20\%                                   & 81.31\%                                  & \cellcolor[HTML]{EFEFEF}79.89\%                     & \cellcolor[HTML]{EFEFEF}82.31\%          & \cellcolor[HTML]{EFEFEF}82.50\%          & 81.54\%                                             & 82.56\%                                  & 82.68\%                                  \\
                            & 5k                                                                             & 84.95\%                                              & 89.33\%                                  & 87.37\%                                  & \cellcolor[HTML]{EFEFEF}86.25\%                     & \cellcolor[HTML]{EFEFEF}89.11\%          & \cellcolor[HTML]{EFEFEF}88.54\%         & 87.75\%                                             & 88.73\%                                  & 88.94\%                                  \\
                            & 10k                                                                            & 87.51\%                                              & 92.05\%                                  & 89.95\%                                  & \cellcolor[HTML]{EFEFEF}88.79\%                     & \cellcolor[HTML]{EFEFEF}92.01\%          & \cellcolor[HTML]{EFEFEF}91.01\%         & 91.07\%                                             & 92.07\%                                  & 92.31\%                                  \\
                            & 20k                                                                            & 88.72\%                                              & 93.46\%                                  & 91.15\%                                  & \cellcolor[HTML]{EFEFEF}90.09\%                     & \cellcolor[HTML]{EFEFEF}93.35\%          & \cellcolor[HTML]{EFEFEF}92.40\%          & 92.80\%                                              & 93.87\%                                  & 94.13\%                                  \\
\multirow{-5}{*}{ResNet-18} & 45k                                                                            & \textbf{89.84\%}                                     & \textbf{95.28\%}                         & \textbf{91.71\%}                         & \cellcolor[HTML]{EFEFEF}\textbf{91.49\%}            & \cellcolor[HTML]{EFEFEF}\textbf{95.29\%} & \cellcolor[HTML]{EFEFEF}\textbf{93.40\%} & \textbf{94.19\%}                                    & \textbf{95.39\%}                         & \textbf{95.44\%}                         \\ \addlinespace
                            & 1k                                                                             & \cellcolor[HTML]{EFEFEF}77.72\%                      & \cellcolor[HTML]{EFEFEF}81.70\%           & \cellcolor[HTML]{EFEFEF}80.01\%          & 76.69\%                                             & 79.35\%                                  & 78.74\%                                 & \cellcolor[HTML]{EFEFEF}80.58\%                     & \cellcolor[HTML]{EFEFEF}81.61\%          & \cellcolor[HTML]{EFEFEF}81.79\%          \\
                            & 5k                                                                             & \cellcolor[HTML]{EFEFEF}83.93\%                      & \cellcolor[HTML]{EFEFEF}88.37\%          & \cellcolor[HTML]{EFEFEF}86.26\%          & 85.32\%                                             & 88.54\%                                  & 87.24\%                                 & \cellcolor[HTML]{EFEFEF}89.47\%                     & \cellcolor[HTML]{EFEFEF}90.29\%          & \cellcolor[HTML]{EFEFEF}90.91\%          \\
                            & 10k                                                                            & \cellcolor[HTML]{EFEFEF}87.60\%                       & \cellcolor[HTML]{EFEFEF}92.60\%           & \cellcolor[HTML]{EFEFEF}89.69\%          & 88.51\%                                             & 91.91\%                                  & 90.51\%                                 & \cellcolor[HTML]{EFEFEF}91.30\%                      & \cellcolor[HTML]{EFEFEF}92.26\%          & \cellcolor[HTML]{EFEFEF}92.73\%          \\
                            & 20k                                                                            & \cellcolor[HTML]{EFEFEF}89.36\%                      & \cellcolor[HTML]{EFEFEF}94.64\%          & \cellcolor[HTML]{EFEFEF}91.32\%          & 90.73\%                                             & 94.73\%                                  & 92.39\%                                 & \cellcolor[HTML]{EFEFEF}93.77\%                     & \cellcolor[HTML]{EFEFEF}94.79\%          & \cellcolor[HTML]{EFEFEF}95.16\%          \\
\multirow{-5}{*}{ResNet-34} & 45k                                                                            & \cellcolor[HTML]{EFEFEF}\textbf{90.35\%}             & \cellcolor[HTML]{EFEFEF}\textbf{96.29\%} & \cellcolor[HTML]{EFEFEF}\textbf{91.95\%} & \textbf{92.15\%}                                    & \textbf{96.67\%}                         & \textbf{93.50\%}                         & \cellcolor[HTML]{EFEFEF}\textbf{95.37\%}            & \cellcolor[HTML]{EFEFEF}\textbf{96.49\%} & \cellcolor[HTML]{EFEFEF}\textbf{96.84\%} \\ \bottomrule
\end{tabular}
\label{tab:attacks_substitute_arch}
\end{table*}

The results presented in \Cref{tab:correct_incorrect} clearly demonstrate that substitute models learn correct predictions significantly better than incorrect predictions. Therefore, if the target model makes fewer mistakes, its behaviour is easier to copy, which is represented by higher fidelity. Moreover, we speculate that higher incorrect prediction accuracy for SimpleNet and pre-trained ResNet-34 models in \Cref{tab:correct_incorrect} is the consequence of using the same training strategy (training from scratch vs transfer learning) for both target and substitute models. We elaborate more on the usage of transfer learning later in \Cref{sec:factor:transfer_learning}. 

\textbf{Conclusion.}
\textit{The target model performance acts as a limiting factor for the fidelity of a substitute model. This behaviour occurs because learning the mistakes of the target model is significantly more challenging compared to the correct predictions. Therefore, for a strong attacker, targeting a better-performing model will lead to a better-performing attack.}

\subsection{Factor 2: Substitute Architecture Choice}
\label{sec:factor:substitute_architecture}
\textbf{Prior Work Discussion.} In more than 10 recent works, the impact of the substitute architecture was studied \cite{papernot_practical_2017, pal_activethief_2020, juuti_prada_2019, orekondy_knockoff_2019, yuan_es_2022, yu_cloudleak_2020, gong_inversenet_2021, zhang_thief_2021, chen_d-dae_2023, yan_holistic_2023, pape_limitations_2023, liu_efficient_2024, karmakar_marich_2023}. While comparing the architectures of substitute and target models, the majority of prior studies concluded that the substitute has to be of the same or higher complexity in order to achieve higher effectiveness. However, some studies encountered the opposite, with more complex models performing worse \cite{zhang_thief_2021, liu_efficient_2024, lin_quda_2023}. In our work, we obtain a similar effect: ResNet-18, in some cases, outperforms the more complex ResNet-34. 

In general, the comparison of target and substitute architectures can be misleading. The target architecture can, for some reason, be selected to have a learning capacity significantly larger than is needed for the classification task. However, selecting a simple architecture that meets the task requirements can be enough for the adversary. 

The substitute model has to learn the task from the data labelled by the target model. Therefore, data should be the key factor in deciding on a substitute architecture. Rather than comparing target and substitute architectures, we suggest measuring how applicable the substitute architecture is for the learning task. This can be done, for example, by evaluating the performance of the substitute model on the original classification task when trained on the original data with the original (truth) labels. 

\textbf{Results.} 
\Cref{tab:attacks_substitute_arch} presents the performance of substitute models with different architectures trained on the original (CIFAR-10) data. As a substitute model, SimpleNet has the lowest scores for all target models and query budgets. However, as a target model, SimpleNet also has the lowest accuracy, suggesting that its learning capacity is likely too limited to learn CIFAR-10 better. 

For the smallest query budgets of 1,000 samples, ResNet-18 outperforms ResNet-34. This trend sometimes persists for up to 20,000 queries. Therefore, having the same architecture as the target model does not imply the best attack performance. ResNet-34 performs the best for 45,000 queries for all target models. As the number of samples grows, the substitute model has to process more information. Consequently, in this scenario, a more complex architecture is beneficial.

We observe similar trends on PD data (CINIC-10): SimpleNet has the lowest performance, and ResNet-34 reaches the highest scores for the largest query budget (see \Cref{tab:attacks_substitute_arch_cinic} in \Cref{appendix:additional_results}). However, unlike for original data, ResNet-18 mostly performs worse than ResNet-34. As we show later in \Cref{sec:factor:data}, CINIC-10 is more complex than CIFAR-10. Hence, it is expected that it requires a more complex architecture such as ResNet-34. 

For the artificial data, on the smaller datasets (1,000 and 5,000 queries), ResNet-18 outperforms ResNet-34 (see \Cref{tab:attacks_substitute_arch_artificial} in \Cref{appendix:additional_results}). As we demonstrate later, the artificial dataset is simpler than CIFAR-10 (\Cref{sec:factor:data}). Hence, especially for small query budgets, a ResNet-18 architecture is complex enough to learn the artificial data.  

\textbf{Conclusion.}
\textit{The complexity of the substitute architecture should correspond to the complexity of the attacker's data rather than be compared to the complexity of the target model. A more complex substitute model should be justified by data characteristics. In particular, selecting a more complex architecture can be beneficial if the adversary gathers more data or uses data of a higher complexity. }

\begin{table*}[t]
\centering
\caption{Performance of substitute models trained on 45,000 samples of the original (CIFAR-10) data.}
\begin{tabular}{lccccccccc}
\toprule

\textbf{Target $\rightarrow$ }   & \multicolumn{3}{c}{\textbf{SimpleNet}}                                                                                                                                     & \multicolumn{3}{c}{\begin{tabular}[c]{@{}c@{}}\textbf{ResNet-34} \\ \textbf{(from scratch)}\end{tabular}}                                                                         & \multicolumn{3}{c}{\begin{tabular}[c]{@{}c@{}}\textbf{ResNet-34} \\ \textbf{(transfer learning)}\end{tabular}}                                                                     \\ \cmidrule(lr){2-4} \cmidrule(lr){5-7} \cmidrule(lr){8-10}
\textbf{Substitute $\downarrow$} & \multicolumn{1}{c}{\textit{Joint Acc}} & \multicolumn{1}{c}{\textit{Acc}}                             & \textit{Fid}                             & \multicolumn{1}{c}{\textit{Joint Acc}} & \multicolumn{1}{c}{\textit{Acc}}                             & \textit{Fid}                            & \multicolumn{1}{c}{\textit{Joint Acc}} & \multicolumn{1}{c}{\textit{Acc}}                             & \textit{Fid}                             \\ \midrule
SimpleNet               & \multicolumn{1}{c}{\cellcolor[HTML]{FFFFFF}87.02\%}                      & \multicolumn{1}{c}{\cellcolor[HTML]{FFFFFF}90.27\%} & \cellcolor[HTML]{FFFFFF}90.61\% & \multicolumn{1}{c}{88.34\%}                                             & \multicolumn{1}{c}{90.45\%}                         & 91.41\%                        & \multicolumn{1}{c}{\cellcolor[HTML]{FFFFFF}89.08\%}                     & \multicolumn{1}{c}{\cellcolor[HTML]{FFFFFF}90.12\%} & \cellcolor[HTML]{FFFFFF}90.33\% \\ 
ResNet-18               & \multicolumn{1}{c}{89.84\%}                                              & \multicolumn{1}{c}{95.28\%}                         & 91.71\%                         & \multicolumn{1}{c}{\cellcolor[HTML]{FFFFFF}91.49\%}                     & \multicolumn{1}{c}{\cellcolor[HTML]{FFFFFF}95.29\%} & \cellcolor[HTML]{FFFFFF}93.40\% & \multicolumn{1}{c}{94.19\%}                                             & \multicolumn{1}{c}{95.39\%}                         & 95.44\%                         \\ 
ResNet-34               & \multicolumn{1}{c}{\cellcolor[HTML]{FFFFFF}90.35\%}                      & \multicolumn{1}{c}{\cellcolor[HTML]{FFFFFF}96.29\%} & \cellcolor[HTML]{FFFFFF}91.95\% & \multicolumn{1}{c}{92.15\%}                                             & \multicolumn{1}{c}{96.67\%}                         & 93.50\%                         & \multicolumn{1}{c}{\cellcolor[HTML]{FFFFFF}95.37\%}                     & \multicolumn{1}{c}{\cellcolor[HTML]{FFFFFF}96.49\%} & \cellcolor[HTML]{FFFFFF}96.84\% \\ \bottomrule
\end{tabular}

\label{tab:attack_substitute_arch_45k}
\end{table*}

\begin{table*}[th]
\centering
\caption{Fidelity scores of SimpleNet substitute model trained on the artificial dataset.}
\begin{tabular}{lcccccc}
\toprule
                                                    & \multicolumn{5}{c}{\textbf{Fidelity}}                                                                                                                & \multirow{2}{*}{\begin{tabular}[c]{@{}c@{}}\textbf{Target accuracy} \\ \textbf{on validation set}\end{tabular}} \\ \cmidrule(lr){2-6}
\multicolumn{1}{l}{\textbf{Target model}}                  & \multicolumn{1}{c}{\textit{1k}}      & \multicolumn{1}{c}{\textit{5k}}      & \multicolumn{1}{c}{\textit{10k}}     & \multicolumn{1}{c}{\textit{20k}}     & \textit{45k}              &                                                                                               \\ \midrule
\multicolumn{1}{l}{SimpleNet}                     & \multicolumn{1}{c}{23.10\%}  & \multicolumn{1}{c}{\textbf{40.77\%}} & \multicolumn{1}{c}{\textbf{48.61\%}} & \multicolumn{1}{c}{\textbf{58.12\%}} & \textbf{68.35\%} & 88.46\%                                                                                       \\ 
\multicolumn{1}{l}{ResNet-34 (from scratch)}      & \multicolumn{1}{c}{\textbf{24.58\%}} & \multicolumn{1}{c}{37.78\%} & \multicolumn{1}{c}{43.49\%} & \multicolumn{1}{c}{54.81\%} & 68.34\% & 91.40\%                                                                                        \\ 
\multicolumn{1}{l}{ResNet-34 (transfer learning)} & \multicolumn{1}{c}{23.57\%} & \multicolumn{1}{c}{35.04\%} & \multicolumn{1}{c}{38.28\%} & \multicolumn{1}{c}{44.95\%} & 51.01\% & 96.78\%                                                                                       \\ \bottomrule
\end{tabular}

\label{tab:attacks_simplenet_sdcifar10}
\end{table*}

\subsection{Factor 3: Usage of Transfer Learning}
\label{sec:factor:transfer_learning}
\textbf{Prior Work Discussion.}
Atli et al. \cite{atli_extraction_2020} demonstrated that stealing models trained from scratch results in lower effectiveness than stealing pre-trained models. In their experiments, all substitute models were trained with transfer learning. Moreover, Zhang et al. \cite{zhang_thief_2021} showed that without pre-trained weights, the effectiveness of the substitute model attacks significantly degrades. However, in their experiments, the dataset used for transfer learning overlapped with the original training data of the target model. Therefore, the pre-trained weights were actually (at least partially) trained on the original data, providing even more advantage for the adversary. We take the previous observations further and demonstrate that substitutes trained from scratch learn better from target models trained from scratch compared to learning from pre-trained target models.

\textbf{Results.} 
In our experiments, two target models were trained from scratch (SimpleNet and ResNet-34), and one was trained with transfer learning from ImageNet (ResNet-34). Further, one substitute model is trained from scratch (SimpleNet), and two are trained from pre-trained on ImageNet weights (ResNet-18 and ResNet-34). \Cref{tab:attack_substitute_arch_45k} illustrates that SimpleNet as a substitute reaches higher fidelity and accuracy scores when targeting models trained from scratch. However, joint accuracy is higher when targeting the pre-trained model, which also has the best performance. This observation, on the one hand, reinforces our findings from \Cref{sec:factor:target_model} that it is easier to learn (i) from better-performing models and (2) correct predictions. On the other hand, higher joint accuracy together with lower fidelity means that SimpleNet can capture mistakes of the pre-trained ResNet-34 worse than mistakes of models trained from scratch. 
For the pre-trained substitute models, all metrics, especially joint accuracy and fidelity, increase with increasing performance of the target model. Even when the target model has a different architecture and is trained from scratch (SimpleNet), stealing it with a pre-trained substitute (ResNet models) yields high scores.

When using artificial data for training substitutes from scratch, the difference between attacks on pre-trained and trained-from-scratch models becomes even more significant. As \Cref{tab:attacks_simplenet_sdcifar10} illustrates, the SimpleNet substitute model trained with artificial data has 17\% lower fidelity when targeting the pre-trained model. For reference, we also report the performance of the target models on the artificial validation set. The trend is the same as with the original data: SimpleNet has the lowest performance, and pre-trained ResNet-34 has the highest. Despite that, the pre-trained target leads to the lowest attack performance.

\textbf{Conclusion.}
\textit{Training target models from scratch does not make them less prone to be stolen. In fact, if a substitute model is trained from scratch, it reaches higher scores when attacking the target model trained from scratch compared to the model trained with transfer learning. Consequently, for the adversary, it is beneficial to follow the same training strategy as the one used for the target model. }

\subsection{Factor 4: Attacker's Data Quality}
\label{sec:factor:data}
\textbf{Prior Work Discussion.}
As shown previously in \Cref{sec:related_work}, PD data has been the least explored in prior works. Correia-Silva et al. \cite{correia-silva_copycat_2018} compared substitutes trained on PD and NPD data. In some settings, models trained on NPD data outperformed ones trained on PD data, leading to the conclusion that random real data is enough for an effective attack. However, in those experiments, the attack with NPD data used significantly more queries than the attack with PD data. In this work, we fix the query budget so that it does not exceed the original training set and explore how PD-like data of different quality and complexity impacts the attack performance.  

\textbf{Results.} 
\Cref{tab:attacks_dataset} demonstrates the performance of substitute models trained with ResNet-18 architecture on different types of attacker's data. The highest scores for all target models are obtained with the original (CIFAR-10) data, followed by PD (CINIC-10) data and PD-like artificial data. Surprisingly, the artificial data labelled by the SimpleNet target model was more beneficial for the attack than the artificial data labelled by ResNet target models, even though the SimpleNet model has the lowest performance on the artificial dataset (see \Cref{tab:attacks_simplenet_sdcifar10}). 

\begin{table*}[h]
\centering
\caption{Performance of ResNet-18 substitute models trained on different attacker's data.}
\begin{tabular}{ccccccccccc}
\toprule
                                                   & \textbf{Target $\rightarrow$}                                                           & \multicolumn{3}{c}{\textbf{SimpleNet}}                                                                                                                                                       & \multicolumn{3}{c}{\begin{tabular}[c]{@{}c@{}}\textbf{ResNet-34} \\ \textbf{(from scratch)}\end{tabular}}                                                                                            & \multicolumn{3}{c}{\begin{tabular}[c]{@{}c@{}}\textbf{ResNet-34} \\ \textbf{(transfer learning)}\end{tabular}}                                                                                       \\ \cmidrule(lr){3-5} \cmidrule(lr){6-8} \cmidrule(lr){9-11}
\multicolumn{1}{c}{\textbf{Dataset $\downarrow$}}         &  \textit{QB} & \multicolumn{1}{c}{\textit{Joint Acc}} & \multicolumn{1}{c}{\textit{Acc}}                                      & \textit{Fid}                                      & \multicolumn{1}{c}{\textit{Joint Acc}} & \multicolumn{1}{c}{\textit{Acc}}                                      & \textit{Fid}                                      & \multicolumn{1}{c}{\textit{Joint Acc}} & \multicolumn{1}{c}{\textit{Acc}}                                      & \textit{Fid}                                      \\ \midrule
\multicolumn{1}{c}{}                             & 1k                                                                             & \multicolumn{1}{c}{\cellcolor[HTML]{EFEFEF}78.54\%}                      & \multicolumn{1}{c}{\cellcolor[HTML]{EFEFEF}82.20\%}           & \cellcolor[HTML]{EFEFEF}81.31\%          & \multicolumn{1}{c}{\cellcolor[HTML]{FFFFFF}79.89\%}                     & \multicolumn{1}{c}{\cellcolor[HTML]{FFFFFF}82.31\%}          & \cellcolor[HTML]{FFFFFF}82.50\%           & \multicolumn{1}{c}{\cellcolor[HTML]{EFEFEF}81.54\%}                     & \multicolumn{1}{c}{\cellcolor[HTML]{EFEFEF}82.56\%}          & \cellcolor[HTML]{EFEFEF}82.68\%          \\  
\multicolumn{1}{c}{}                             & 5k                                                                             & \multicolumn{1}{c}{\cellcolor[HTML]{EFEFEF}84.95\%}                      & \multicolumn{1}{c}{\cellcolor[HTML]{EFEFEF}89.33\%}          & \cellcolor[HTML]{EFEFEF}87.37\%          & \multicolumn{1}{c}{\cellcolor[HTML]{FFFFFF}86.25\%}                     & \multicolumn{1}{c}{\cellcolor[HTML]{FFFFFF}89.11\%}          & \cellcolor[HTML]{FFFFFF}88.54\%          & \multicolumn{1}{c}{\cellcolor[HTML]{EFEFEF}87.75\%}                     & \multicolumn{1}{c}{\cellcolor[HTML]{EFEFEF}88.73\%}          & \cellcolor[HTML]{EFEFEF}88.94\%          \\ 
\multicolumn{1}{c}{}                             & 10k                                                                            & \multicolumn{1}{c}{\cellcolor[HTML]{EFEFEF}87.51\%}                      & \multicolumn{1}{c}{\cellcolor[HTML]{EFEFEF}92.05\%}          & \cellcolor[HTML]{EFEFEF}89.95\%          & \multicolumn{1}{c}{\cellcolor[HTML]{FFFFFF}88.79\%}                     & \multicolumn{1}{c}{\cellcolor[HTML]{FFFFFF}92.01\%}          & \cellcolor[HTML]{FFFFFF}91.01\%          & \multicolumn{1}{c}{\cellcolor[HTML]{EFEFEF}91.07\%}                     & \multicolumn{1}{c}{\cellcolor[HTML]{EFEFEF}92.07\%}          & \cellcolor[HTML]{EFEFEF}92.31\%          \\ 
\multicolumn{1}{c}{}                             & 20k                                                                            & \multicolumn{1}{c}{\cellcolor[HTML]{EFEFEF}88.72\%}                      & \multicolumn{1}{c}{\cellcolor[HTML]{EFEFEF}93.46\%}          & \cellcolor[HTML]{EFEFEF}91.15\%          & \multicolumn{1}{c}{\cellcolor[HTML]{FFFFFF}90.09\%}                     & \multicolumn{1}{c}{\cellcolor[HTML]{FFFFFF}93.35\%}          & \cellcolor[HTML]{FFFFFF}92.40\%           & \multicolumn{1}{c}{\cellcolor[HTML]{EFEFEF}92.80\%}                      & \multicolumn{1}{c}{\cellcolor[HTML]{EFEFEF}93.87\%}          & \cellcolor[HTML]{EFEFEF}94.13\%          \\ 
\multicolumn{1}{c}{\multirow{-5}{*}{CIFAR-10}}    & 45k                                                                            & \multicolumn{1}{c}{\cellcolor[HTML]{EFEFEF}\textbf{89.84\%}}             & \multicolumn{1}{c}{\cellcolor[HTML]{EFEFEF}\textbf{95.28\%}} & \cellcolor[HTML]{EFEFEF}\textbf{91.71\%} & \multicolumn{1}{c}{\cellcolor[HTML]{FFFFFF}\textbf{91.49\%}}            & \multicolumn{1}{c}{\cellcolor[HTML]{FFFFFF}\textbf{95.29\%}} & \cellcolor[HTML]{FFFFFF}\textbf{93.40\%}  & \multicolumn{1}{c}{\cellcolor[HTML]{EFEFEF}\textbf{94.19\%}}            & \multicolumn{1}{c}{\cellcolor[HTML]{EFEFEF}\textbf{95.39\%}} & \cellcolor[HTML]{EFEFEF}\textbf{95.44\%} \\ \addlinespace
\multicolumn{1}{c}{}                             & 1k                                                                             & \multicolumn{1}{c}{65.65\%}                                              & \multicolumn{1}{c}{68.60\%}                                  & 68.76\%                                  & \multicolumn{1}{c}{\cellcolor[HTML]{EFEFEF}69.14\%}                     & \multicolumn{1}{c}{\cellcolor[HTML]{EFEFEF}71.32\%}          & \cellcolor[HTML]{EFEFEF}71.49\%          & \multicolumn{1}{c}{72.19\%}                                             & \multicolumn{1}{c}{73.20\%}                                   & 73.41\%                                  \\  
\multicolumn{1}{c}{}                             & 5k                                                                             & \multicolumn{1}{c}{76.33\%}                                              & \multicolumn{1}{c}{79.60\%}                                   & 79.41\%                                  & \multicolumn{1}{c}{\cellcolor[HTML]{EFEFEF}79.79\%}                     & \multicolumn{1}{c}{\cellcolor[HTML]{EFEFEF}82.19\%}          & \cellcolor[HTML]{EFEFEF}82.24\%          & \multicolumn{1}{c}{82.78\%}                                             & \multicolumn{1}{c}{83.81\%}                                  & 84.06\%                                  \\  
\multicolumn{1}{c}{}                             & 10k                                                                            & \multicolumn{1}{c}{80.10\%}                                               & \multicolumn{1}{c}{83.48\%}                                  & 83.23\%                                  & \multicolumn{1}{c}{\cellcolor[HTML]{EFEFEF}82.25\%}                     & \multicolumn{1}{c}{\cellcolor[HTML]{EFEFEF}84.83\%}          & \cellcolor[HTML]{EFEFEF}84.69\%          & \multicolumn{1}{c}{85.09\%}                                             & \multicolumn{1}{c}{86.13\%}                                  & 86.38\%                                  \\  
\multicolumn{1}{c}{}                             & 20k                                                                            & \multicolumn{1}{c}{81.91\%}                                              & \multicolumn{1}{c}{85.21\%}                                  & 85.40\%                                   & \multicolumn{1}{c}{\cellcolor[HTML]{EFEFEF}84.84\%}                     & \multicolumn{1}{c}{\cellcolor[HTML]{EFEFEF}87.32\%}          & \cellcolor[HTML]{EFEFEF}87.52\%          & \multicolumn{1}{c}{87.90\%}                                              & \multicolumn{1}{c}{89.01\%}                                  & 89.12\%                                  \\  
\multicolumn{1}{c}{\multirow{-5}{*}{CINIC-10}}    & 45k                                                                            & \multicolumn{1}{c}{\textbf{84.28\%}}                                     & \multicolumn{1}{c}{\textbf{87.59\%}}                         & \textbf{87.83\%}                         & \multicolumn{1}{c}{\cellcolor[HTML]{EFEFEF}\textbf{86.88\%}}            & \multicolumn{1}{c}{\cellcolor[HTML]{EFEFEF}\textbf{89.43\%}} & \cellcolor[HTML]{EFEFEF}\textbf{89.64\%} & \multicolumn{1}{c}{\textbf{90.42\%}}                                    & \multicolumn{1}{c}{\textbf{91.44\%}}                         & \textbf{91.76\%}                         \\ \addlinespace
\multicolumn{1}{c}{}                             & 1k                                                                             & \multicolumn{1}{c}{\cellcolor[HTML]{EFEFEF}63.28\%}                      & \multicolumn{1}{c}{\cellcolor[HTML]{EFEFEF}66.20\%}           & \cellcolor[HTML]{EFEFEF}65.86\%          & \multicolumn{1}{c}{64.13\%}                                             & \multicolumn{1}{c}{66.06\%}                                  & 66.32\%                                  & \multicolumn{1}{c}{\cellcolor[HTML]{EFEFEF}66.05\%}                     & \multicolumn{1}{c}{\cellcolor[HTML]{EFEFEF}66.83\%}          & \cellcolor[HTML]{EFEFEF}67.10\%           \\  
\multicolumn{1}{c}{}                             & 5k                                                                             & \multicolumn{1}{c}{\cellcolor[HTML]{EFEFEF}67.90\%}                       & \multicolumn{1}{c}{\cellcolor[HTML]{EFEFEF}71.11\%}          & \cellcolor[HTML]{EFEFEF}70.55            & \multicolumn{1}{c}{67.30\%}                                              & \multicolumn{1}{c}{69.44\%}                                  & 69.52\%                                  & \multicolumn{1}{c}{\cellcolor[HTML]{EFEFEF}71.12\%}                     & \multicolumn{1}{c}{\cellcolor[HTML]{EFEFEF}71.97\%}          & \cellcolor[HTML]{EFEFEF}72.14\%          \\  
\multicolumn{1}{c}{}                             & 10k                                                                            & \multicolumn{1}{c}{\cellcolor[HTML]{EFEFEF}69.93\%}                      & \multicolumn{1}{c}{\cellcolor[HTML]{EFEFEF}73.04\%}          & \cellcolor[HTML]{EFEFEF}72.86\%          & \multicolumn{1}{c}{71.38\%}                                             & \multicolumn{1}{c}{73.72\%}                                  & 73.40\%                                   & \multicolumn{1}{c}{\cellcolor[HTML]{EFEFEF}69.58\%}                     & \multicolumn{1}{c}{\cellcolor[HTML]{EFEFEF}70.46\%}          & \cellcolor[HTML]{EFEFEF}70.63\%          \\  
\multicolumn{1}{c}{}                             & 20k                                                                            & \multicolumn{1}{c}{\cellcolor[HTML]{EFEFEF}73.43\%}                      & \multicolumn{1}{c}{\cellcolor[HTML]{EFEFEF}76.72\%}          & \cellcolor[HTML]{EFEFEF}76.42\%          & \multicolumn{1}{c}{73.96\%}                                             & \multicolumn{1}{c}{76.31\%}                                  & 76.19\%                                  & \multicolumn{1}{c}{\cellcolor[HTML]{EFEFEF}68.08\%}                     & \multicolumn{1}{c}{\cellcolor[HTML]{EFEFEF}69.01\%}          & \cellcolor[HTML]{EFEFEF}69.16\%          \\  
\multicolumn{1}{c}{\multirow{-5}{*}{Artificial}} & 45k                                                                            & \multicolumn{1}{c}{\cellcolor[HTML]{EFEFEF}\textbf{75.69\%}}             & \multicolumn{1}{c}{\cellcolor[HTML]{EFEFEF}\textbf{78.88\%}} & \cellcolor[HTML]{EFEFEF}\textbf{78.78\%} & \multicolumn{1}{c}{\textbf{74.94\%}}                                    & \multicolumn{1}{c}{\textbf{77.10\%}}                          & \textbf{77.28\%}                         & \multicolumn{1}{c}{\cellcolor[HTML]{EFEFEF}\textbf{72.96\%}}            & \multicolumn{1}{c}{\cellcolor[HTML]{EFEFEF}\textbf{73.81\%}} & \cellcolor[HTML]{EFEFEF}\textbf{74.00\%}    \\ \bottomrule
\end{tabular}

\label{tab:attacks_dataset}
\end{table*}

\begin{table}[th]
\centering
\caption{Comparison of validation and test scores of attacks using CINIC-10 dataset.}

\begin{tabular}{p{1.6cm}ccccc}
\toprule
                                                                                                             &                                                                          & \multicolumn{2}{c}{\textbf{Validation scores}}                                           & \multicolumn{2}{c}{\textbf{Test Scores}}                                                 \\ \cmidrule(lr){3-4} \cmidrule(lr){5-6} 
\multirow{-2}{*}{\textbf{Target model}}                                                                               & \multirow{-2}{*}{\begin{tabular}[c]{@{}c@{}}\textbf{Query}\\ \textbf{budget}\end{tabular}} & \multicolumn{1}{c}{\textit{Acc}}                                      & \textit{Fid}              & \multicolumn{1}{c}{\textit{Acc}}                                      & \textit{Fid}              \\ \midrule
                                                                                                             & 1k                                                                       & \multicolumn{1}{c}{61.22\%}                                  & 66.28\%          & \multicolumn{1}{c}{68.60\%}                                  & 68.76\%          \\  
                                                                                                             & 5k                                                                       & \multicolumn{1}{c}{65.51\%}                                  & 70.34\%          & \multicolumn{1}{c}{79.60\%}                                   & 79.41\%          \\  
                                                                                                             & 10k                                                                      & \multicolumn{1}{c}{66.99\%}                                  & 72.20\%           & \multicolumn{1}{c}{83.48\%}                                  & 83.23\%          \\  
                                                                                                             & 20k                                                                      & \multicolumn{1}{c}{66.49\%}                                  & 74.01\%          & \multicolumn{1}{c}{85.21\%}                                  & 85.40\%           \\  
\multirow{-5}{*}{SimpleNet}                                                                                  & 45k                                                                      & \multicolumn{1}{c}{\textbf{67.24\%}}                         & \textbf{76.02\%} & \multicolumn{1}{c}{\textbf{87.59\%}}                         & \textbf{87.83\%} \\ \addlinespace
\rowcolor[HTML]{FFFFFF} 
\cellcolor[HTML]{FFFFFF}                                                                                     & 1k                                                                       & \multicolumn{1}{c}{\cellcolor[HTML]{FFFFFF}64.30\%}           & 67.70\%           & \multicolumn{1}{c}{\cellcolor[HTML]{FFFFFF}71.32\%}          & 71.49\%          \\  
\rowcolor[HTML]{FFFFFF} 
\cellcolor[HTML]{FFFFFF}                                                                                     & 5k                                                                       & \multicolumn{1}{c}{\cellcolor[HTML]{FFFFFF}68.36\%}          & 72.16\%          & \multicolumn{1}{c}{\cellcolor[HTML]{FFFFFF}82.19\%}          & 82.24\%          \\ 
\rowcolor[HTML]{FFFFFF} 
\cellcolor[HTML]{FFFFFF}                                                                                     & 10k                                                                      & \multicolumn{1}{c}{\cellcolor[HTML]{FFFFFF}70.25\%}          & 74.13\%          & \multicolumn{1}{c}{\cellcolor[HTML]{FFFFFF}84.83\%}          & 84.69\%          \\ 
\rowcolor[HTML]{FFFFFF} 
\cellcolor[HTML]{FFFFFF}                                                                                     & 20k                                                                      & \multicolumn{1}{c}{\cellcolor[HTML]{FFFFFF}70.23\%}          & 75.89\%          & \multicolumn{1}{c}{\cellcolor[HTML]{FFFFFF}87.32\%}          & 87.52\%          \\  
\rowcolor[HTML]{FFFFFF} 
\multirow{-5}{*}{\cellcolor[HTML]{FFFFFF}\begin{tabular}[l]{@{}l@{}}ResNet-34\\ (from scratch)\end{tabular}} & 45k                                                                      & \multicolumn{1}{c}{\cellcolor[HTML]{FFFFFF}\textbf{71.02\%}} & \textbf{77.88\%} & \multicolumn{1}{c}{\cellcolor[HTML]{FFFFFF}\textbf{89.43\%}} & \textbf{89.64\%} \\ \addlinespace
                                                                                                             & 1k                                                                       & \multicolumn{1}{c}{67.16\%}                                  & 72.18\%          & \multicolumn{1}{c}{73.20\%}                                   & 73.41\%          \\  
                                                                                                             & 5k                                                                       & \multicolumn{1}{c}{73.22\%}                                  & 78.65\%          & \multicolumn{1}{c}{83.81\%}                                  & 84.06\%          \\  
                                                                                                             & 10k                                                                      & \multicolumn{1}{c}{74.83\%}                                  & 80.73\%          & \multicolumn{1}{c}{86.13\%}                                  & 86.38\%          \\  
                                                                                                             & 20k                                                                      & \multicolumn{1}{c}{75.79\%}                                  & 81.83\%          & \multicolumn{1}{c}{89.01\%}                                  & 89.12\%          \\  
\multirow{-5}{*}{\begin{tabular}[l]{@{}l@{}}ResNet-34\\ (transfer learning)\end{tabular}}                    & 45k                                                                      & \multicolumn{1}{c}{\textbf{77.50\%}}                          & \textbf{83.48\%} & \multicolumn{1}{c}{\textbf{91.44\%}}                         & \textbf{91.76\%} \\ \bottomrule
\end{tabular}

\label{tab:attacks_cinic10_val_test}
\end{table}

For further analysis, we compare scores obtained on the attacker's validation set with the CIFAR-10 test scores. This comparison shows the difference between the performance estimation of the attack by the adversary and the eventual real success of the attack. For the CIFAR-10 attacker's dataset, validation and test scores are very similar, as both datasets are drawn from the same distribution. 

For CINIC-10 (shown in \Cref{tab:attacks_cinic10_val_test}), the performance on the attacker's validation set is notably lower than on the CIFAR-10 test set, indicating that CINIC-10 is more complex. Hence, using CINIC-10, an adversary tends to underestimate the power of their attacks. In contrast, the scores measured on the artificial validation set are way higher than the test scores (see \Cref{tab:attacks_sdcifar10_val_test}). 
For example, with a query budget of 1,000, all substitute models reached an accuracy larger than 93\% on the validation set. With the same query budget, substitutes trained on the CIFAR-10 data (\Cref{tab:attacks_baseline}) reached at most 82\% of accuracy on the validation set. This further indicates that the artificial data is not complex enough, and its quality has to be improved. Using artificial data, an adversary will likely overestimate the performance of the substitute model on the real data.

As \Cref{tab:attacks_dataset} shows, attacks using the CINIC-10 dataset, which we concluded is more complex than the original data, in general follow the same trends as attacks using the original data and have reasonable performance. 
Therefore, we conclude that data that is more feasible for attacks should have either comparable complexity or even be more complex than the original data, which can be assessed by comparing validation and test scores. However, as we only consider data with the same context as the original data, this statement needs further investigation for non-problem domain attacks. 
\begin{table}[t]
\centering
\caption{Comparison of validation and test scores of attacks using artificial dataset.}
\begin{tabular}{p{1.6cm}ccccc}
\toprule
                                                                                                             &                                                                          & \multicolumn{2}{c}{\textbf{Validation scores}}                                           & \multicolumn{2}{c}{\textbf{Test Scores}}                                                \\ \cmidrule(lr){3-4} \cmidrule(lr){5-6} 
\multirow{-2}{*}{\textbf{Target model}}                                                                               & \multirow{-2}{*}{\begin{tabular}[c]{@{}c@{}}\textbf{Query}\\ \textbf{budget}\end{tabular}} & \multicolumn{1}{c}{\textit{Acc}}                                      & \textit{Fid}              & \multicolumn{1}{c}{\textit{Acc}}                                     & \textit{Fid}              \\ \midrule
                                                                                                             & 1k                                                                       & \multicolumn{1}{c}{93.18\%}                                  & 88.66\%          & \multicolumn{1}{c}{66.20\%}                                  & 65.86\%          \\  
                                                                                                             & 5k                                                                       & \multicolumn{1}{c}{\textbf{94.26\%}}                         & 89.76\%          & \multicolumn{1}{c}{71.11\%}                                 & 70.55\%          \\  
                                                                                                             & 10k                                                                      & \multicolumn{1}{c}{93.46\%}                                  & 89.82\%          & \multicolumn{1}{c}{73.04\%}                                 & 72.86\%          \\  
                                                                                                             & 20k                                                                      & \multicolumn{1}{c}{93.58\%}                                  & 90.22\%          & \multicolumn{1}{c}{76.72\%}                                 & 76.42\%          \\  
\multirow{-5}{*}{SimpleNet}                                                                                  & 45k                                                                      & \multicolumn{1}{c}{93.34\%}                                  & \textbf{90.80\%}  & \multicolumn{1}{c}{\textbf{78.88\%}}                        & \textbf{78.78\%} \\ \addlinespace
\rowcolor[HTML]{FFFFFF} 
\cellcolor[HTML]{FFFFFF}                                                                                     & 1k                                                                       & \multicolumn{1}{c}{\cellcolor[HTML]{FFFFFF}94.58\%}          & 91.22\%          & \multicolumn{1}{c}{\cellcolor[HTML]{FFFFFF}66.06\%}         & 66.32\%          \\  
\rowcolor[HTML]{FFFFFF} 
\cellcolor[HTML]{FFFFFF}                                                                                     & 5k                                                                       & \multicolumn{1}{c}{\cellcolor[HTML]{FFFFFF}\textbf{95.74\%}} & 92.24\%          & \multicolumn{1}{c}{\cellcolor[HTML]{FFFFFF}69.44\%}         & 69.52\%          \\ 
\rowcolor[HTML]{FFFFFF} 
\cellcolor[HTML]{FFFFFF}                                                                                     & 10k                                                                      & \multicolumn{1}{c}{\cellcolor[HTML]{FFFFFF}95.06\%}          & 92.38\%          & \multicolumn{1}{c}{\cellcolor[HTML]{FFFFFF}73.72\%}         & 73.40\%           \\  
\rowcolor[HTML]{FFFFFF} 
\cellcolor[HTML]{FFFFFF}                                                                                     & 20k                                                                      & \multicolumn{1}{c}{\cellcolor[HTML]{FFFFFF}95.58\%}          & 92.84\%          & \multicolumn{1}{c}{\cellcolor[HTML]{FFFFFF}76.31\%}         & 76.19\%          \\  
\rowcolor[HTML]{FFFFFF} 
\multirow{-5}{*}{\cellcolor[HTML]{FFFFFF}\begin{tabular}[l]{@{}l@{}}ResNet-34\\ (from scratch)\end{tabular}} & 45k                                                                      & \multicolumn{1}{c}{\cellcolor[HTML]{FFFFFF}95.08\%}          & \textbf{92.94\%} & \multicolumn{1}{c}{\cellcolor[HTML]{FFFFFF}\textbf{77.10\%}} & \textbf{77.28\%} \\ \addlinespace
                                                                                                             & 1k                                                                       & \multicolumn{1}{c}{95.94\%}                                  & 95.50\%           & \multicolumn{1}{c}{66.83\%}                                 & 67.10\%           \\ 
                                                                                                             & 5k                                                                       & \multicolumn{1}{c}{96.92\%}                                  & 96.46\%          & \multicolumn{1}{c}{71.97\%}                                 & 72.14\%          \\  
                                                                                                             & 10k                                                                      & \multicolumn{1}{c}{97.38\%}                                  & 97.10\%           & \multicolumn{1}{c}{70.46\%}                                 & 70.63\%          \\  
                                                                                                             & 20k                                                                      & \multicolumn{1}{c}{97.58\%}                                  & 97.10\%           & \multicolumn{1}{c}{69.01\%}                                 & 69.16\%          \\  
\multirow{-5}{*}{\begin{tabular}[l]{@{}l@{}}ResNet-34\\ (transfer learning)\end{tabular}}                    & 45k                                                                      & \multicolumn{1}{c}{\textbf{97.74\%}}                         & \textbf{97.50\%}  & \multicolumn{1}{c}{\textbf{73.81\%}}                        & \textbf{74.00\%}    \\ \bottomrule
\end{tabular}

\label{tab:attacks_sdcifar10_val_test}
\end{table}

\textbf{Conclusion.}
\textit{PD data more complex than the original training data leads to better-performing substitute models than simpler PD data. Furthermore, based on substitute validation scores, attackers with more complex data tend to underestimate the performance of the substitute model. In turn, having data simpler than the original leads to an overestimation of substitute performance.}

\subsection{Factor 5: Attacker's Capabilities}
\textbf{Prior Work Discussion.}
As presented earlier in \Cref{tab:substitute_attacks_comparison}, query optimisation techniques have been widely studied in related work. Prior results suggest that with optimisation, an attack can achieve higher scores with the same number of queries \cite{pal_activethief_2020, yu_cloudleak_2020}. Similarly, query optimisation can help to reach the same attack performance level while using fewer queries. However, as we observed earlier in \Cref{sec:related_work}, query optimisation has not been applied for data-free attacks. In earlier work, these attacks were usually carried out using millions of queries and are the least efficient substitute training attacks (see \Cref{sec:comparison_sota}). While our data-free attack already has high efficiency, we demonstrate that it can be improved with active learning.

\begin{table*}[pb]
\centering
\small
\caption{Performance of ResNet-18 substitute models trained on the artificial data with different query optimisation techniques. In parentheses, we report the difference with the non-optimised attack.}
\begin{tabular}{cccccccc}
\toprule
\multirow{2}{*}{\begin{tabular}[c]{@{}c@{}}\textbf{Query} \\ \textbf{optimisation $\downarrow$}\end{tabular}}                                                                                              & \textbf{Target $\rightarrow$} & \multicolumn{2}{c}{\textbf{SimpleNet}}                                                                                            & \multicolumn{2}{c}{\textbf{ResNet-34  (from scratch)}}                                  & \multicolumn{2}{c}{\textbf{ResNet-34  (transfer learning)}}                             \\ \cmidrule(lr){3-4} \cmidrule(lr){5-6} \cmidrule(lr){7-8}
            & \textit{QB}                   & \textit{Acc}                                                    & \textit{Fid}                                                    & \textit{Acc}                                                    & \textit{Fid}                                                    & \textit{Acc}                                                    & \textit{Fid}                                                    \\ \midrule
                                                                                              & 1k                            & \cellcolor[HTML]{EFEFEF}{\color[HTML]{9A0000} 64.62\% \scriptsize{(-1.58\%)}} & \cellcolor[HTML]{EFEFEF}{\color[HTML]{9A0000} 64.09\% \scriptsize{(-1.77\%)}} & {\color[HTML]{036400} 68.30\% \scriptsize{(+2.24\%)}}                         & {\color[HTML]{036400} 68.24\% \scriptsize{(+1.92\%)}}                         & \cellcolor[HTML]{EFEFEF}{\color[HTML]{036400} 68.02\% \scriptsize{(+1.19\%)}} & \cellcolor[HTML]{EFEFEF}{\color[HTML]{036400} 68.18\% \scriptsize{(+1.08\%)}} \\
                                                                                              & 5k                            & \cellcolor[HTML]{EFEFEF}{\color[HTML]{9A0000} 70.65\% \scriptsize{(-0.46\%)}} & \cellcolor[HTML]{EFEFEF}{\color[HTML]{9A0000} 70.39\% \scriptsize{(-0.16\%)}} & {\color[HTML]{036400} 72.75\% \scriptsize{(+3.31\%)}}                         & {\color[HTML]{036400} 72.66\% \scriptsize{(+3.14\%)}}                         & \cellcolor[HTML]{EFEFEF}{\color[HTML]{9A0000} 69.86\% \scriptsize{(-2.11\%)}} & \cellcolor[HTML]{EFEFEF}{\color[HTML]{9A0000} 69.99\% \scriptsize{(-2.15\%)}} \\
                                                                                              & 10k                           & \cellcolor[HTML]{EFEFEF}{\color[HTML]{036400} 76.64\% \scriptsize{(+3.60\%)}} & \cellcolor[HTML]{EFEFEF}{\color[HTML]{036400} 76.14\% \scriptsize{(+3.28\%)}} & {\color[HTML]{036400} 78.51\% \scriptsize{(+4.79\%)}}                         & {\color[HTML]{036400} 78.55\% \scriptsize{(+5.15\%)}}                         & \cellcolor[HTML]{EFEFEF}{\color[HTML]{036400} 75.39\% \scriptsize{(+4.93\%)}} & \cellcolor[HTML]{EFEFEF}{\color[HTML]{036400} 75.61\% \scriptsize{(+4.98\%)}} \\
\multirow{-4}{*}{\begin{tabular}[c]{@{}c@{}}Active\\ learning\end{tabular}}                   & 20k                           & \cellcolor[HTML]{EFEFEF}{\color[HTML]{036400} 77.98\% \scriptsize{(+1.26\%)}} & \cellcolor[HTML]{EFEFEF}{\color[HTML]{036400} 77.65\% \scriptsize{(+1.23\%)}} & {\color[HTML]{036400} 78.91\% \scriptsize{(+2.60\%)}}                         & {\color[HTML]{036400} 78.79\% \scriptsize{(+2.60\%)}}                         & \cellcolor[HTML]{EFEFEF}{\color[HTML]{036400} 73.47\% \scriptsize{(+4.46\%)}} & \cellcolor[HTML]{EFEFEF}{\color[HTML]{036400} 73.62\% \scriptsize{(+4.46\%)}} \\ \addlinespace
                                                                                              & 1k                            & {\color[HTML]{9A0000} 63.60\% \scriptsize{(-2.60\%)}}                         & {\color[HTML]{9A0000} 63.03\% \scriptsize{(-2.83\%)}}                         & \cellcolor[HTML]{EFEFEF}{\color[HTML]{9A0000} 65.69\% \scriptsize{(-0.37\%)}} & \cellcolor[HTML]{EFEFEF}{\color[HTML]{9A0000} 65.83\% \scriptsize{(-0.49\%)}} & {\color[HTML]{9A0000} 64.45\% \scriptsize{(-2.38\%)}}                         & {\color[HTML]{9A0000} 64.78\% \scriptsize{(-2.32\%)}}                         \\
                                                                                              & 5k                            & {\color[HTML]{9A0000} 68.16\% \scriptsize{(-2.95\%)}}                         & {\color[HTML]{9A0000} 67.53\% \scriptsize{(-3.02\%)}}                         & \cellcolor[HTML]{EFEFEF}{\color[HTML]{036400} 70.75\% \scriptsize{(+1.31\%)}} & \cellcolor[HTML]{EFEFEF}{\color[HTML]{036400} 70.46\% \scriptsize{(+0.94\%)}} & {\color[HTML]{9A0000} 68.84\% \scriptsize{(-3.13\%)}}                         & {\color[HTML]{9A0000} 68.92\% \scriptsize{(-3.22\%)}}                         \\
                                                                                              & 10k                           & {\color[HTML]{9A0000} 69.47\% \scriptsize{(-3.57\%)}}                         & {\color[HTML]{9A0000} 69.07\% \scriptsize{(-3.79\%)}}                         & \cellcolor[HTML]{EFEFEF}{\color[HTML]{036400} 76.16\% \scriptsize{(+2.44\%)}} & \cellcolor[HTML]{EFEFEF}{\color[HTML]{036400} 76.24\% \scriptsize{(+2.84\%)}} & {\color[HTML]{9A0000} 70.28\% \scriptsize{(-0.18\%)}}                         & {\color[HTML]{9A0000} 70.44\% \scriptsize{(-0.19\%)}}                         \\
                                                                                              & 20k                           & {\color[HTML]{9A0000} 75.39\% \scriptsize{(-1.33\%)}}                         & {\color[HTML]{9A0000} 75.26\% \scriptsize{(-1.16\%)}}                         & \cellcolor[HTML]{EFEFEF}{\color[HTML]{9A0000} 76.29\% \scriptsize{(-0.02\%)}} & \cellcolor[HTML]{EFEFEF}{\color[HTML]{036400} 76.48\% \scriptsize{(+0.29\%)}} & {\color[HTML]{036400} 72.04\% \scriptsize{(+3.03\%)}}                         & {\color[HTML]{036400} 72.14\% \scriptsize{(+2.98\%)}}                         \\
\multirow{-5}{*}{\begin{tabular}[c]{@{}c@{}}Adversarial\\ augmentation\end{tabular}}          & 45k                           & {\color[HTML]{9A0000} 78.05\% \scriptsize{(-0.83\%)}}                         & {\color[HTML]{9A0000} 77.73\% \scriptsize{(-1.05\%)}}                         & \cellcolor[HTML]{EFEFEF}{\color[HTML]{036400} 77.96\% \scriptsize{(+0.86\%)}} & \cellcolor[HTML]{EFEFEF}{\color[HTML]{036400} 78.18\% \scriptsize{(+0.90\%)}} & {\color[HTML]{036400} 74.39\% \scriptsize{(+0.58\%)}}                         & {\color[HTML]{036400} 74.54\% \scriptsize{(+0.54\%)}}                         \\ \addlinespace
                                                                                              & 1k                            & \cellcolor[HTML]{EFEFEF}{\color[HTML]{9A0000} 63.53\% \scriptsize{(-2.67\%)}} & \cellcolor[HTML]{EFEFEF}{\color[HTML]{9A0000} 63.03\% \scriptsize{(-2.83\%)}} & {\color[HTML]{9A0000} 65.71\% \scriptsize{(-0.35\%)}}                         & {\color[HTML]{9A0000} 65.44\% \scriptsize{(-0.88\%)}}                         & \cellcolor[HTML]{EFEFEF}{\color[HTML]{9A0000} 64.23\% \scriptsize{(-2.60\%)}} & \cellcolor[HTML]{EFEFEF}{\color[HTML]{9A0000} 64.49\% \scriptsize{(-2.61\%)}} \\
                                                                                              & 5k                            & \cellcolor[HTML]{EFEFEF}{\color[HTML]{9A0000} 68.58\% \scriptsize{(-2.53\%)}} & \cellcolor[HTML]{EFEFEF}{\color[HTML]{9A0000} 67.73\% \scriptsize{(-2.82\%)}} & {\color[HTML]{036400} 74.34\% \scriptsize{(+4.90\%)}}                         & {\color[HTML]{036400} 74.20\% \scriptsize{(+4.68\%)}}                         & \cellcolor[HTML]{EFEFEF}{\color[HTML]{9A0000} 69.39\% \scriptsize{(-2.58\%)}} & \cellcolor[HTML]{EFEFEF}{\color[HTML]{9A0000} 69.45\% \scriptsize{(-2.69\%)}} \\
                                                                                              & 10k                           & \cellcolor[HTML]{EFEFEF}{\color[HTML]{036400} 73.84\% \scriptsize{(+0.80\%)}} & \cellcolor[HTML]{EFEFEF}{\color[HTML]{036400} 73.71\% \scriptsize{(+0.85\%)}} & {\color[HTML]{036400} 75.70\% \scriptsize{(+1.98\%)}}                         & {\color[HTML]{036400} 75.72\% \scriptsize{(+2.32\%)}}                         & \cellcolor[HTML]{EFEFEF}{\color[HTML]{036400} 74.57\% \scriptsize{(+4.11\%)}} & \cellcolor[HTML]{EFEFEF}{\color[HTML]{036400} 74.86\% \scriptsize{(+4.23\%)}} \\
                                                                                              & 20k                           & \cellcolor[HTML]{EFEFEF}{\color[HTML]{9A0000} 74.84\% \scriptsize{(-1.88\%)}} & \cellcolor[HTML]{EFEFEF}{\color[HTML]{9A0000} 74.62\% \scriptsize{(-1.80\%)}} & {\color[HTML]{036400} 77.90\% \scriptsize{(+1.59\%)}}                         & {\color[HTML]{036400} 77.88\% \scriptsize{(+1.69\%)}}                         & \cellcolor[HTML]{EFEFEF}{\color[HTML]{036400} 74.41\% \scriptsize{(+5.40\%)}} & \cellcolor[HTML]{EFEFEF}{\color[HTML]{036400} 74.74\% \scriptsize{(+5.58\%)}} \\
\multirow{-5}{*}{\begin{tabular}[c]{@{}c@{}}Active\\ adversarial\\ augmentation\end{tabular}} & 45k                           & \cellcolor[HTML]{EFEFEF}{\color[HTML]{036400} 79.42\% \scriptsize{(+0.54\%)}} & \cellcolor[HTML]{EFEFEF}{\color[HTML]{036400} 79.16\% \scriptsize{(+0.38\%)}} & {\color[HTML]{036400} 77.41\% \scriptsize{(+0.31\%)}}                         & {\color[HTML]{036400} 77.58\% \scriptsize{(+0.30\%)}}                         & \cellcolor[HTML]{EFEFEF}{\color[HTML]{036400} 75.50\% \scriptsize{(+1.69\%)}}  & \cellcolor[HTML]{EFEFEF}{\color[HTML]{036400} 75.71\% \scriptsize{(+1.71\%)}} \\ \bottomrule
\end{tabular}

\label{tab:attacks_optimisation}
\end{table*}

\textbf{Results.} 
\Cref{tab:attacks_optimisation} shows the performance scores of the attacks optimised for the scenario of the weakest attacker in terms of knowledge about the target model and its data. All attacks in the table are performed with a ResNet-18 substitute model trained on the artificial dataset. 
For each score in \Cref{tab:attacks_optimisation}, we also provide the difference to the scores from  \Cref{tab:attacks_dataset}, which contains attacks with the same settings but without any optimisation techniques. Therefore, a positive number means the optimisation technique improved the performance, while negative numbers indicate a degradation of the attack's performance. There is no reported active learning attack with 45,000 queries, as that is the size of the whole dataset, and applying active learning does not change the final attacker's set compared to the non-optimised attack.

In most of the configurations, active learning performs the best, although for the smaller query budgets (1,000 and 5,000), it does not always improve the baseline, and in a few cases, other optimisation techniques reach higher scores. However, active learning improves the attack and holds the second-best score in all such cases. For small query budgets, the initial seed (set of samples on which the model is trained before the first optimisation round, see \Cref{appendix:query_optimisation_methods}) might be too small for the substitute model to learn how to rank the most valuable samples. 
In all configurations with a query budget of 10,000 or more, there is an optimisation technique that improves the baseline attack. Hence, applying those techniques, especially active learning, is most useful for mid-range query budgets.  

With active learning, our data-free attack reaches more than 75\% of fidelity and accuracy with a query budget of only 10,000 queries. The state-of-the-art results presented in \Cref{sec:comparison_sota}, are, while having similar effectiveness performance, three orders of magnitude less efficient. Therefore, we are the first to demonstrate that even with the weakest attacker's knowledge, substitute training attacks can be effectively performed with less data than the target model was trained on. 

\textbf{Conclusion.}
\textit{Query optimisation techniques are the most beneficial for mid-range query budgets starting from 10,000 queries. Smaller query budgets (1,000 and 5,000) are not enough for a substitute model to learn how to rank the most valuable samples.}

\textit{Data-free attacks on image classifiers can be effective with query budgets smaller than the original training set. Weak attacker's knowledge does not require to be overcompensated with significantly increased querying capabilities.}

\section{Comparison with the State-of-the-Art}
\label{sec:comparison_sota}
We conclude our attack analysis by comparing the performance of our attacks with papers that performed model stealing attacks against CIFAR-10 classifiers. We summarise the performance scores in \Cref{tab:attacks_vs_sota_below_40} for query budgets below 40,000 and in \Cref{tab:attacks_vs_sota_45+} for query budgets from 45,000. 
Additionally, we provide information about the attacker's data, target model outputs, and the performance of the target model on the CIFAR-10 test set. For attack effectiveness evaluation, we report accuracy and fidelity of substitute models. Efficiency is represented by the exact number of queries used by an attack and \textit{query score}. The latter shows how many queries are needed per a target model's training sample, i.e. it is a relation between the attacker's dataset and the CIFAR-10 training dataset (50,000 samples). If some information is not reported by a paper, we mark it as \unknown. Some values are estimated based on the information provided in the papers (see \Cref{appendix:value_estimation_for_sota} for more details). For our work, we report scores for attacks using query optimisation, ResNet-18 as a substitute model, and ResNet-34 trained from scratch as the target model.

\begin{table*}
\centering
\caption{Comparison of attacks implemented in this work with the state-of-the-art for query budgets below 40,000.}
\begin{tabular}{llllrrrrl}
\toprule
\textbf{Query budget} & \textbf{Data}        & \textbf{Outputs}           & \textbf{Paper}                                 & \textbf{Target Acc.} & \textbf{Sub. Acc.} & \textbf{Sub. Fid.} & \textbf{Queries}   & \textbf{Query score} \\ \midrule
\multirow{8}{*}{$<2.5$k}                               & Original    & \unknown          & \cite{pengcheng_query-efficient_2018} & $>$91\%                                               &                                                     & 77.47\%                                             & 1.6k      & 0.032                                                                          \\
                                                       & Original    & \unknown          & \cite{zhao_extracting_2023}           & \unknown                                              &                                                     & 76.40\%                                             & 2k        & 0.04                                                                           \\
                                                       & Original    & Gradients         & \cite{milli_model_2019}               & est. 90\%                                             & est. 88\%                                           &                                                     & 1k        & 0.02                                                                           \\
                                                       & Original    & Probabilities     & \cite{jagielski_high_2020}            & 95.75\%                                               & 90.63\%                                             & 91.39\%                                             & 1k        & 0.02                                                                           \\
                                                       & Original    & Probabilities     & \cite{he_drmi_2021}                   & 93.90\%                                               & 82.28\%                                             &                                                     & 1k        & 0.02                                                                           \\
                                                       & Original    & Labels            & this work                             & 93.61\%                                               & 84.51\%                                             & 84.23\%                                             & 1k        & 0.02                                                                           \\
                                                       & PD          & Labels            & this work                             & 93.61\%                                               & 71.58\%                                             & 71.99\%                                             & 1k        & 0.02                                                                           \\
                                                       & Artificial  & Labels            & this work                             & 93.61\%                                               & 68.30\%                                             & 68.24\%                                             & 1k        & 0.02                                                                           \\ \cmidrule(lr){1-9}
\multirow{13}{*}{$<10$k}                               & Original    & \unknown          & \cite{pengcheng_query-efficient_2018} & $>$91\%                                               &                                                     & 77.96\%                                             & 6.4k      & 0.128                                                                          \\
                                                       & Original    & \unknown          & \cite{zhao_extracting_2023}           & \unknown                                              &                                                     & 80.25\%                                             & 4k        & 0.08                                                                           \\
                                                       & Original    & Probabilities+XAI & \cite{yan_towards_2022}               & 92.03\%                                               & est. 77\%                                           &                                                     & 8k        & 0.16                                                                           \\
                                                       & Original    & Probabilities     & \cite{jagielski_high_2020}            & 95.75\%                                               & 93.29\%                                             & 93.99\%                                             & 4k        & 0.08                                                                           \\
                                                       & Original    & Probabilities     & \cite{he_drmi_2021}                   & 93.90\%                                               & 89.74\%                                             &                                                     & 4k        & 0.08                                                                           \\
                                                                  & Original    & Labels+XAI        & \cite{yan_explanation_2023}           & 92.03\%                                               &                                                     & 72.50\%                                              & 8k        & 0.16                                                                           \\
                                                       & Original    & Labels            & \cite{yan_holistic_2023}              & 90.45\%                                               & 58.95\%                                             & 59.90\%                                             & 8k        & 0.16                                                                           \\
                                                       & Original    & Labels            & this work                             & 93.61\%                                               & 90.22\%                                             & 89.20\%                                             & 5k        & 0.1                                                                            \\
                                                       & PD          & Labels            & this work                             & 93.61\%                                               & 82.96\%                                             & 83.03\%                                             & 5k        & 0.1                                                                            \\
                                                       & NPD         & Labels            & \cite{yan_holistic_2023}              & 90.45\%                                               & 31.21\%                                             & 32.20\%                                             & 8k        & 0.16                                                                           \\
                                                       & NPD         & Labels            & \cite{karmakar_marich_2023}           & 91.82\%                                               & 71.65\%                                             &                                                     & 8.4k      & 0.168                                                                          \\
                                                       & NPD         & Labels            & \cite{yang_swifttheft_2024}         & \unknown                                              &                                                     & 71.45\%                                             & 8k        & 0.16                                                                           \\
                                                       & Artificial  & Labels            & this work                             & 93.61\%                                               & 74.34\%                                             & 74.20\%                                             & 5k        & 0.1                                                                            \\ \cmidrule(lr){1-9}
\multirow{11}{*}{$<20$k}                               & Original    & \unknown          & \cite{pengcheng_query-efficient_2018} & $>$91\%                                               &                                                     & 83.61\%                                             & 12.8k     & 0.256                                                                          \\
                                                       & Original    & Probabilities     & \cite{he_drmi_2021}                   & 93.90\%                                               & 92.50\%                                             &                                                     & 10k       & 0.2                                                                            \\
                                                       & Original    & Labels            & this work                             & 93.61\%                                               & 93.05\%                                             & 91.93\%                                             & 10k       & 0.2                                                                            \\
                                                       & PD          & Labels            & this work                             & 93.61\%                                               & 85.85\%                                             & 85.95\%                                             & 10k       & 0.2                                                                            \\
                                                       & NPD         & Probabilities     & \cite{pal_activethief_2020}           & \unknown                                              &                                                     & 77.29\%                                             & 10k       & 0.2                                                                            \\
                                                       & NPD         & Probabilities     & \cite{gong_inversenet_2021}           & \unknown                                              &                                                     & 77.80\%                                             & 10k       & 0.2                                                                            \\
                                                       & NPD         & Probabilities     & \cite{khaled_careful_2022}            & \unknown                                              & est. 69\%                                           & est. 68\%                                           & 15k       & 0.3                                                                            \\
                                                       & NPD         & Labels            & \cite{pal_activethief_2020}           & \unknown                                              &                                                     & 64.23\%                                             & 10k       & 0.2                                                                            \\
                                                       & NPD         & Labels            & \cite{gong_inversenet_2021}           & \unknown                                              &                                                     & 75.40\%                                             & 10k       & 0.2                                                                            \\
                                                       & NPD         & Labels            & \cite{yang_swifttheft_2024}         & \unknown                                              & \textbf{}                                           & 74.06\%                                             & 12k       & 0.24                                                                           \\
                                                       & Artificial  & Labels            & this work                             & 93.61\%                                               & 78.51\%                                             & 78.55\%                                             & 10k       & 0.2                                                                            \\ \cmidrule(lr){1-9}
\multirow{11}{*}{$<40$k}                               & Original    & \unknown          & \cite{pengcheng_query-efficient_2018} & $>$91\%                                               &                                                     & 84.12\%                                             & 25.6k     & 0.512                                                                          \\
                                                       & Original    & Labels            & \cite{pape_limitations_2023}          & 93.70\%                                               &                                                     & 94.09\%                                             & est. 25k  & 0.5                                                                            \\
                                                       & Original    & Labels            & this work                             & 93.61\%                                               & 94.61\%                                             & 92.89\%                                             & 20k       & 0.4                                                                            \\
                                                       & PD          & Labels            & this work                             & 93.61\%                                               & 88.21\%                                             & 88.69\%                                             & 20k       & 0.4                                                                            \\
                                                       & NPD         & Probabilities     & \cite{khaled_careful_2022}            & \unknown                                              & est. 71\%                                           & est. 70\%                                           & 30k       & 0.6                                                                            \\
                                                       & NPD         & Labels            & \cite{pal_activethief_2020}           & \unknown                                              &                                                     & 78.36\%                                             & 30k       & 0.6                                                                            \\
                                                       & NPD         & Labels            & \cite{wang_enhance_2022}              & \unknown                                              & 80.90\%                                             &                                                     & 30k       & 0.6                                                                            \\
                                                       & NPD         & Labels            & \cite{wang_black-box_2022}            & 91.56\%                                               & 80.47\%                                             & 82.14\%                                             & 30k       & 0.6                                                                            \\
                                                       & NPD         & Labels            & \cite{jindal_army_2024}        & 92.18\%                                               & 83.06\%                                             & 84.12\%                                             & 30k       & 0.6                                                                            \\
                                                       & NPD         & Labels            & \cite{yang_swifttheft_2024}         & \unknown                                              &                                                     & 80.67\%                                             & 20k       & 0.4                                                                            \\
                                                       & Artificial  & Labels            & this work                             & 93.61\%                                               & 78.91\%                                             & 78.79\%                                             & 20k       & 0.4                                                                            \\ \bottomrule
\end{tabular}
\label{tab:attacks_vs_sota_below_40}
\end{table*}

\begin{table*}
\centering
\caption{Comparison of attacks implemented in this work with the state-of-the-art for query budgets from 45,000.}
\begin{tabular}{lllrrrrr}
\toprule
\textbf{Data}                & \textbf{Outputs}  & \textbf{Paper}                    & \textbf{Target Acc.} & \textbf{Sub. Acc.} & \textbf{Sub. Fid.} & \textbf{Queries} & \textbf{Query score} \\ \midrule
Original                     & Labels            & this work                         & 93.61\%              & 95.29\%            & 93.40\%            & 45k              & 0.9                  \\ \cmidrule(lr){1-8}
\multirow{2}{*}{PD}          & Probabilities     & \cite{correia-silva_copycat_2018} & 95.30\%              & 90.00\%            &                    & 269k             & 5.38                 \\ \addlinespace
                             & Labels            & this work                         & 93.61\%              & 89.71\%            & 90.03\%            & 45k              & 0.9                  \\ \cmidrule(lr){1-8}
\multirow{8}{*}{NPD}         & Probabilities     & \cite{khaled_careful_2022}        & \unknown             & est. 76\%          & est. 75\%          & 50k              & 1                    \\
                             & Probabilities     & \cite{correia-silva_copycat_2018} & 95.30\%              & 94.00\%            &                    & 3.4m             & 68                   \\
                             & Probabilities     & \cite{atli_extraction_2020}       & 94.60\%              & 88.20\%            &                    & 100k             & 2                    \\
                             & Probabilities     & \cite{barbalau_black-box_2020}    & 82.50\%              & 79.00\%            &                    & est. $>$50k      & est. $>$1            \\ \addlinespace
                             & Labels            & \cite{mosafi_stealing_2019}       & 90.48\%              & 89.59\%            &                    & est. $>$1m       & est. $>$20           \\
                             & Labels            & \cite{pal_activethief_2020}       & \unknown             &                    & 81.57\%            & 100k             & 2                    \\
                             & Labels            & \cite{pal_activethief_2020}       & \unknown             &                    & 84.99\%            & 120k             & 2.4                  \\
                             & Labels            & \cite{atli_extraction_2020}       & 94.60\%              & 53.60\%            &                    & 100k             & 2                    \\ \cmidrule(lr){1-8}
\multirow{23}{*}{Artificial} & \unknown          & \cite{zhang_towards_2022}         & \unknown             & 61.90\%            &                    & 60k              & 1.2                  \\ \addlinespace
                             & Probabilities+XAI & \cite{miura_megex_2024}           & 95.50\%              & 72.10\%            &                    & 1m               & 20                   \\
                             & Probabilities+XAI & \cite{miura_megex_2024}           & 95.50\%              & 90.40\%            &                    & 10m              & 200                  \\
                             & Probabilities+XAI & \cite{miura_megex_2024}           & 95.50\%              & 92.30\%            &                    & 20m              & 400                  \\\addlinespace
                             & Probabilities     & \cite{kariyappa_maze_2021}        & 92.26\%              & 89.85\%            &                    & 30m              & 600                  \\
                             & Probabilities     & \cite{truong_data-free_2021}      & 95.50\%              & 88.10\%            &                    & 20m              & 400                  \\
                             & Probabilities     & \cite{truong_data-free_2021}      & 95.50\%              & 89.90\%            &                    & 30m              & 600                  \\
                             & Probabilities     & \cite{sanyal_towards_2022}        & 95.50\%              & 91.24\%            &                    & 8m               & 160                  \\
                             & Probabilities     & \cite{yuan_es_2022}               & 91.93\%              & 80.79\%            &                    & est. $>$1m       & est. $>$20           \\
                             & Probabilities     & \cite{liu_efficient_2024}        & 93.00\%              & 88.50\%            & 92.10\%            & 20m              & 400                  \\
                             & Probabilities     & \cite{beetham_dual_2023}          & 95.50\%              & 91.34\%            &                    & 20m              & 400                  \\
                             & Probabilities     & \cite{rosenthal_disguide_2023}    & 95.54\%              & 94.02\%            &                    & 20m              & 400                  \\
                             & Probabilities     & \cite{hong_exploring_2023}        & 90.71\%              & 84.70\%            &                    & 13m              & 260                  \\
                             & Probabilities     & \cite{lin_quda_2023}              & 81.74\%              & 70.30\%            &                    & 100k             & 2                    \\ \addlinespace
                             & Labels+XAI & \cite{yan_explanation-based_2023} & 95.54\%              & est. 90\%          &                    & 20m              & 400                  \\ \addlinespace
                             & Labels            & \cite{sanyal_towards_2022}        & 95.50\%              & 84.51\%            &                    & 8m               & 160                  \\
                             & Labels            & \cite{yuan_es_2022}               & 91.93\%              & 69.64\%            &                    & est. $>$1m       & est. $>$20           \\
                             & Labels            & \cite{beetham_dual_2023}          & 95.50\%              & 78.72\%            &                    & 20m              & 400                  \\
                             & Labels            & \cite{hong_exploring_2023}        & 90.71\%              & 29.58\%            &                    & 6m               & 120                  \\
                             & Labels            & \cite{rosenthal_disguide_2023}    & 95.54\%              & 87.93\%            &                    & 8m               & 160                  \\
                             & Labels            & \cite{yang_efficient_2023}        & 82.50\%              & 73.10\%            &                    & 1.6m             & 32                   \\
                             & Labels            & \cite{yang_efficient_2023}        & 82.50\%              & 73.81\%            &                    & 17.6m            & 352                  \\
                             & Labels            & this work                         & 93.61\%              & 78.52\%            & 79.06\%            & 45k              & 0.9                 \\ \bottomrule
\end{tabular}
\label{tab:attacks_vs_sota_45+}
\end{table*}

We group attacks with similar query budgets in \Cref{tab:attacks_vs_sota_below_40} (shown in the first column), the same data type (second column), and target outputs (third column). In general, as attacks from the same group have the most similar attacker's strength, a performance comparison should only be launched within a group. However, since often there is no work matching ours, we also compare our attacks with ones having stronger attacker profiles.
Below, we summarise our main findings from \Cref{tab:attacks_vs_sota_below_40} for each query budget.

\textit{Query budget $<$2.5k.} All prior attacks with this query budget relied on availability of original data. Therefore, we are the first one to demonstrate feasibility of attacks with this query budget for PD and artificial data. Our attack with original data is outperformed by two prior works, both having a stronger assumption about the target outputs, namely probabilities \cite{jagielski_high_2020} and gradients \cite{milli_model_2019}. 
    
\textit{Query budget $<$10k.} From a total of seven prior attacks using original data, only one achieves higher performance than ours \cite{jagielski_high_2020}. However, it relies on probabilities whereas we only use labels. Further, even our PD attack outperforms five of prior works using original data. Moreover, our data-free attack outperforms all previous NPD attacks, while using less queries (5,000 compared to 8,000). As earlier, we are the first to use PD or artificial data with such query budget.

\textit{Query budget $<$20k.} Among attacks using original data, our has the highest accuracy and fidelity scores. As was the case before, none of the prior studies used PD or artificial data. However, using artificial data, we achieve higher accuracy and fidelity than all six previous NPD attacks. 

\textit{Query budget $<$40k.} One of the prior attacks using original data has slightly better fidelity than ours \cite{pape_limitations_2023}. Besides, the limitations artificial data become more visible compared to NPD data---most of the prior works outperform our attack, although the difference in performance remains rather small (up to 5.5\%). There are still no other PD or data-free attacks within this query budget to compare our work with.

Notably, we are the first to demonstrate feasibility of data-free attacks with query budgets below 40,000. Besides, for the same query budgets, none of the prior works launched attacks using PD data. Meanwhile, our PD attacks outperform some prior works that use original data, more queries, and more revealing target outputs. Furthermore, our data-free attack has better performance than any NPD attack for query budgets below 20,000. 

We compare the last group of attacks that used at least 45,000 queries in \Cref{tab:attacks_vs_sota_45+}. No prior works used that many queries for attacks based on the original data. For PD data, the only prior work \cite{correia-silva_copycat_2018} marginally outperforms our attack, while using significantly more queries and probabilities as target outputs. 

The most significant group of papers with large query budgets uses artificial data for training a substitute model. Among six data-free attack that use labels as target outputs, our attack is marginally (by 0.2\%) outperformed by one prior work \cite{beetham_dual_2023} and more significantly (by 6-9.5\%) by two other works \cite{sanyal_towards_2022, rosenthal_disguide_2023}. Both of them used 8 million of queries, which is 160 times more than the size of the target model training set. Meanwhile, we only utilised 45,000 queries, proposing the most efficient data-free attack as of now. 

Overall, our data-free attack shows promising results for future work. As optimising artificial data quality was not the prime goal of this work, we expect to reach higher scores with further research. 


\section{Discussion} \label{sec:discussion}
In this section, we discuss two additional aspects important for attack evaluation, namely the transferability of a substitute model and defences against model stealing.

\subsection{Transferability}
In this work, we focused on accuracy and fidelity as performance metrics. However, transferability can be very insightful in estimating how similar two models behave close to the decision boundary. The main difficulty with this metric is that there is no unified way to measure it. Related work uses different adversarial example crafting methods and various hyperparameters defining the strength of the perturbation. For this reason, we omitted to report transferability in \Cref{sec:comparison_sota}, as the scores would simply be incomparable. 

However, as two of our query optimisation techniques utilise adversarial examples, we additionally measured transferability to see if these optimisations could improve the scores. For this purpose, we used exactly the same Deepfool algorithm as in adversarial augmentation  \cite{pengcheng_query-efficient_2018}. However, the results were poor, varying from 0\% to 16\%. We speculate that the reason is that the perturbations Deepfool makes are too minute to impact the decision boundary and improve the transferability significantly. Using a method with stronger perturbations would likely result in a better score.
However, it has to be considered that while it is likely easier to achieve a high transferability score if the adversarial examples are created with stronger perturbation, a commonly agreed upper bound of acceptable perturbation should be established to enable the above-mentioned comparability.

\subsection{Defences against Model Stealing}
Defences against model stealing can be categorised into proactive and reactive \cite{oliynyk_i_2023}. Most common proactive defences are based on adding noise to model inputs or outputs to perturb the information obtained from the target model and mislead the substitute training process. In this work, we only assume labels as outputs, perturbing which will directly lead to target model performance degradation on its original classification task (unlike confidence score perturbations). In safety-critical settings and when the quality of service is crucial, this approach is not feasible. For this reason, we evaluated attacks against defences that incur an acceptable loss of correctness and perturb only a small fraction of predictions. While an acceptable value depends on the specific scenario, we set approximately 1\% of the predicted labels to be modified.  
As a result, outputs of unprotected and protected models were identical for approximately 99\% of data. We tested input perturbation defence by Grana \cite{grana_perturbing_2020} and a region-based classifier by Cao and Gong \cite{cao_mitigating_2017} as an output perturbation (output perturbation defences designed against model stealing assume confidence scores as outputs \cite{orekondy_prediction_2020, chen_das-ast_2021, mazeika_how_2022}). These defences have shown to be ineffective---in most of the cases, the performance of the substitute model did not decrease by more than 1\%. 

Another type of proactive defence is (re-)training the target model from scratch \cite{oliynyk_i_2023} instead of building on a pre-trained model. However, as we showed in \Cref{sec:factor:transfer_learning}, training from scratch also does not protect the target models. Therefore, we conclude that none of the current proactive defences can defend against attacks on image classifiers that use only labels. 

Reactive defences include ownership verification methods (watermarking, fingerprinting) and monitors. While ownership verification methods might be effective against our attacks, they can not protect a model from being stolen and can only serve as evidence. Monitors, which aim to detect an ongoing attack by tracking suspicious queries, should, in most settings, not mark problem-domain-like data as suspicious, as it can lead to false positives, harming API clients. In our work, we use original, problem domain, and problem-domain-like data generated by a diffusion model. Therefore, none of these data distributions should be marked as malicious.

\section{Conclusion} \label{sec:conclusion}
In this paper, we presented new insights into the influence different factors have on the success rate of substitute training attacks. In particular, we demonstrated that attacks benefit from (i) targeting better-performing models, (ii) adopting an architecture that fits the quality and quantity of the attacker's data, (iii) using the target model's training strategy, (iv) having more complex data, and (v) optimising queries, in particular, for data-free attacks. 
While most of these findings suggest that attacks can be improved by some modifications, our result about the target model performance implies that previous attacks could have been underestimated and might produce even higher scores when targeting better models. 

A key takeaway from our work is that attackers can do better---even with less effort. We adopted a simple diffusion-based data generation approach, performed fair hyperparameter tuning on every step of the attack and boosted the performance with query optimisation techniques. As a result, we attained a data-free attack with a query budget of 10,000 queries and performance comparable with those using millions of queries. With original or problem-domain data, we outperform the state-of-the-art, even having weaker assumptions about model outputs while achieving a better efficiency score. \textit{Finally, if attackers indeed can do better---can defenders do better as well?}

\newpage
\section*{Acknowledgement}
The financial support by the Austrian Federal Ministry of Labour and Economy, the National Foundation for Research, Technology and Development and the Christian Doppler Research Association is gratefully acknowledged.

SBA Research (SBA-K1 NGC) is a COMET Centre within the COMET – Competence Centers for Excellent Technologies Programme and funded by BMK, BMAW, and the federal state of Vienna. COMET is managed by FFG.

This research was further funded by the European Union under contract no. 101136305. The Hungarian partner is funded by the Hungarian National Research, Development, and Innovation Fund. Views and opinions expressed are however those of the author(s) only and do not necessarily reflect those of the European Union or the Hungarian National Research, Development and Innovation Fund. Neither the European Union nor the Hungarian National Research, Development and Innovation Fund can be held responsible for them.

\bibliographystyle{IEEEtran}
\bibliography{references}

\begin{thebibliography}{10}
\providecommand{\url}[1]{#1}
\csname url@samestyle\endcsname
\providecommand{\newblock}{\relax}
\providecommand{\bibinfo}[2]{#2}
\providecommand{\BIBentrySTDinterwordspacing}{\spaceskip=0pt\relax}
\providecommand{\BIBentryALTinterwordstretchfactor}{4}
\providecommand{\BIBentryALTinterwordspacing}{\spaceskip=\fontdimen2\font plus
\BIBentryALTinterwordstretchfactor\fontdimen3\font minus \fontdimen4\font\relax}
\providecommand{\BIBforeignlanguage}[2]{{%
\expandafter\ifx\csname l@#1\endcsname\relax
\typeout{** WARNING: IEEEtran.bst: No hyphenation pattern has been}%
\typeout{** loaded for the language `#1'. Using the pattern for}%
\typeout{** the default language instead.}%
\else
\language=\csname l@#1\endcsname
\fi
#2}}
\providecommand{\BIBdecl}{\relax}
\BIBdecl

\bibitem{tramer_stealing_2016}
\BIBentryALTinterwordspacing
F.~Tramèr, F.~Zhang, A.~Juels, M.~K. Reiter, and T.~Ristenpart, ``Stealing {Machine} {Learning} {Models} via {Prediction} {APIs},'' in \emph{25th {USENIX} {Security} {Symposium} ({USENIX} {Security} 16)}.\hskip 1em plus 0.5em minus 0.4em\relax Austin, TX: USENIX Association, Aug. 2016, pp. 601--618. [Online]. Available: \url{https://www.usenix.org/conference/usenixsecurity16/technical-sessions/presentation/tramer}
\BIBentrySTDinterwordspacing

\bibitem{orekondy_knockoff_2019}
\BIBentryALTinterwordspacing
T.~Orekondy, B.~Schiele, and M.~Fritz, ``Knockoff {Nets}: {Stealing} {Functionality} of {Black}-{Box} {Models},'' in \emph{2019 {IEEE}/{CVF} {Conference} on {Computer} {Vision} and {Pattern} {Recognition} ({CVPR})}.\hskip 1em plus 0.5em minus 0.4em\relax Long Beach, CA, USA: IEEE, Jun. 2019, pp. 4949--4958. [Online]. Available: \url{https://ieeexplore.ieee.org/document/8953839/}
\BIBentrySTDinterwordspacing

\bibitem{szyller_good_2021}
\BIBentryALTinterwordspacing
S.~Szyller, V.~Duddu, T.~Gröndahl, and N.~Asokan, ``Good {Artists} {Copy}, {Great} {Artists} {Steal}: {Model} {Extraction} {Attacks} {Against} {Image} {Translation} {Models},'' 2021, version Number: 2. [Online]. Available: \url{https://arxiv.org/abs/2104.12623}
\BIBentrySTDinterwordspacing

\bibitem{krishna_thieves_2020}
K.~Krishna, G.~S. Tomar, A.~P. Parikh, N.~Papernot, and M.~Iyyer, ``Thieves on {Sesame} {Street}! {Model} {Extraction} of {BERT}-based {APIs},'' in \emph{International {Conference} on {Learning} {Representations}}, 2020.

\bibitem{chen_stealing_2021}
\BIBentryALTinterwordspacing
K.~Chen, S.~Guo, T.~Zhang, X.~Xie, and Y.~Liu, ``\BIBforeignlanguage{en}{Stealing {Deep} {Reinforcement} {Learning} {Models} for {Fun} and {Profit}},'' in \emph{\BIBforeignlanguage{en}{Proceedings of the 2021 {ACM} {Asia} {Conference} on {Computer} and {Communications} {Security}}}.\hskip 1em plus 0.5em minus 0.4em\relax Virtual Event Hong Kong: ACM, May 2021, pp. 307--319. [Online]. Available: \url{https://dl.acm.org/doi/10.1145/3433210.3453090}
\BIBentrySTDinterwordspacing

\bibitem{oliynyk_i_2023}
\BIBentryALTinterwordspacing
D.~Oliynyk, R.~Mayer, and A.~Rauber, ``\BIBforeignlanguage{en}{I {Know} {What} {You} {Trained} {Last} {Summer}: {A} {Survey} on {Stealing} {Machine} {Learning} {Models} and {Defences}},'' \emph{\BIBforeignlanguage{en}{ACM Computing Surveys}}, vol.~55, no. 14s, pp. 1--41, Dec. 2023. [Online]. Available: \url{https://dl.acm.org/doi/10.1145/3595292}
\BIBentrySTDinterwordspacing

\bibitem{jagielski_high_2020}
\BIBentryALTinterwordspacing
M.~Jagielski, N.~Carlini, D.~Berthelot, A.~Kurakin, and N.~Papernot, ``High {Accuracy} and {High} {Fidelity} {Extraction} of {Neural} {Networks},'' in \emph{29th {USENIX} {Security} {Symposium} ({USENIX} {Security} 20)}.\hskip 1em plus 0.5em minus 0.4em\relax USENIX Association, Aug. 2020, pp. 1345--1362. [Online]. Available: \url{https://www.usenix.org/conference/usenixsecurity20/presentation/jagielski}
\BIBentrySTDinterwordspacing

\bibitem{papernot_practical_2017}
\BIBentryALTinterwordspacing
N.~Papernot, P.~McDaniel, I.~Goodfellow, S.~Jha, Z.~B. Celik, and A.~Swami, ``\BIBforeignlanguage{en}{Practical {Black}-{Box} {Attacks} against {Machine} {Learning}},'' in \emph{\BIBforeignlanguage{en}{Proceedings of the 2017 {ACM} on {Asia} {Conference} on {Computer} and {Communications} {Security}}}.\hskip 1em plus 0.5em minus 0.4em\relax Abu Dhabi United Arab Emirates: ACM, Apr. 2017, pp. 506--519. [Online]. Available: \url{https://dl.acm.org/doi/10.1145/3052973.3053009}
\BIBentrySTDinterwordspacing

\bibitem{correia-silva_copycat_2018}
\BIBentryALTinterwordspacing
J.~R. Correia-Silva, R.~F. Berriel, C.~Badue, A.~F. De~Souza, and T.~Oliveira-Santos, ``Copycat {CNN}: {Stealing} {Knowledge} by {Persuading} {Confession} with {Random} {Non}-{Labeled} {Data},'' in \emph{2018 {International} {Joint} {Conference} on {Neural} {Networks} ({IJCNN})}.\hskip 1em plus 0.5em minus 0.4em\relax Rio de Janeiro: IEEE, Jul. 2018, pp. 1--8. [Online]. Available: \url{https://ieeexplore.ieee.org/document/8489592/}
\BIBentrySTDinterwordspacing

\bibitem{pal_activethief_2020}
\BIBentryALTinterwordspacing
S.~Pal, Y.~Gupta, A.~Shukla, A.~Kanade, S.~Shevade, and V.~Ganapathy, ``{ActiveThief}: {Model} {Extraction} {Using} {Active} {Learning} and {Unannotated} {Public} {Data},'' \emph{Proceedings of the AAAI Conference on Artificial Intelligence}, vol.~34, no.~01, pp. 865--872, Apr. 2020. [Online]. Available: \url{https://ojs.aaai.org/index.php/AAAI/article/view/5432}
\BIBentrySTDinterwordspacing

\bibitem{juuti_prada_2019}
\BIBentryALTinterwordspacing
M.~Juuti, S.~Szyller, S.~Marchal, and N.~Asokan, ``{PRADA}: {Protecting} {Against} {DNN} {Model} {Stealing} {Attacks},'' in \emph{2019 {IEEE} {European} {Symposium} on {Security} and {Privacy} ({EuroS}\&{P})}.\hskip 1em plus 0.5em minus 0.4em\relax Stockholm, Sweden: IEEE, Jun. 2019, pp. 512--527. [Online]. Available: \url{https://ieeexplore.ieee.org/document/8806737/}
\BIBentrySTDinterwordspacing

\bibitem{atli_extraction_2020}
B.~G. Atli, S.~Szyller, M.~Juuti, S.~Marchal, and N.~Asokan, ``Extraction of {Complex} {DNN} {Models}: {Real} {Threat} or {Boogeyman}?'' in \emph{Engineering {Dependable} and {Secure} {Machine} {Learning} {Systems}}, O.~Shehory, E.~Farchi, and G.~Barash, Eds.\hskip 1em plus 0.5em minus 0.4em\relax Cham: Springer International Publishing, 2020, pp. 42--57.

\bibitem{pengcheng_query-efficient_2018}
\BIBentryALTinterwordspacing
L.~Pengcheng, J.~Yi, and L.~Zhang, ``Query-{Efficient} {Black}-{Box} {Attack} by {Active} {Learning},'' in \emph{2018 {IEEE} {International} {Conference} on {Data} {Mining} ({ICDM})}.\hskip 1em plus 0.5em minus 0.4em\relax Singapore: IEEE, Nov. 2018, pp. 1200--1205. [Online]. Available: \url{https://ieeexplore.ieee.org/document/8594968/}
\BIBentrySTDinterwordspacing

\bibitem{mosafi_stealing_2019}
\BIBentryALTinterwordspacing
I.~Mosafi, E.~O. David, and N.~S. Netanyahu, ``Stealing {Knowledge} from {Protected} {Deep} {Neural} {Networks} {Using} {Composite} {Unlabeled} {Data},'' in \emph{2019 {International} {Joint} {Conference} on {Neural} {Networks} ({IJCNN})}.\hskip 1em plus 0.5em minus 0.4em\relax Budapest, Hungary: IEEE, Jul. 2019, pp. 1--8. [Online]. Available: \url{https://ieeexplore.ieee.org/document/8851798/}
\BIBentrySTDinterwordspacing

\bibitem{yuan_es_2022}
\BIBentryALTinterwordspacing
X.~Yuan, L.~Ding, L.~Zhang, X.~Li, and D.~O. Wu, ``{ES} {Attack}: {Model} {Stealing} {Against} {Deep} {Neural} {Networks} {Without} {Data} {Hurdles},'' \emph{IEEE Transactions on Emerging Topics in Computational Intelligence}, vol.~6, no.~5, pp. 1258--1270, Oct. 2022. [Online]. Available: \url{https://ieeexplore.ieee.org/document/9726514/}
\BIBentrySTDinterwordspacing

\bibitem{milli_model_2019}
\BIBentryALTinterwordspacing
S.~Milli, L.~Schmidt, A.~D. Dragan, and M.~Hardt, ``\BIBforeignlanguage{en}{Model {Reconstruction} from {Model} {Explanations}},'' in \emph{\BIBforeignlanguage{en}{Proceedings of the {Conference} on {Fairness}, {Accountability}, and {Transparency}}}.\hskip 1em plus 0.5em minus 0.4em\relax Atlanta GA USA: ACM, Jan. 2019, pp. 1--9. [Online]. Available: \url{https://dl.acm.org/doi/10.1145/3287560.3287562}
\BIBentrySTDinterwordspacing

\bibitem{kariyappa_maze_2021}
S.~Kariyappa, A.~Prakash, and M.~K. Qureshi, ``{MAZE}: {Data}-{Free} {Model} {Stealing} {Attack} {Using} {Zeroth}-{Order} {Gradient} {Estimation},'' in \emph{Proceedings of the {IEEE}/{CVF} {Conference} on {Computer} {Vision} and {Pattern} {Recognition} ({CVPR})}, Jun. 2021, pp. 13\,814--13\,823.

\bibitem{roberts_model_2019}
\BIBentryALTinterwordspacing
N.~Roberts, V.~U. Prabhu, and M.~McAteer, ``Model {Weight} {Theft} {With} {Just} {Noise} {Inputs}: {The} {Curious} {Case} of the {Petulant} {Attacker},'' in \emph{{ICML} {Workshop} on the {Security} and {Privacy} of {Machine} {Learning}}, Long Beach, CA, Jun. 2019. [Online]. Available: \url{https://icml2019workshop.github.io/}
\BIBentrySTDinterwordspacing

\bibitem{barbalau_black-box_2020}
A.~Bărbălău, A.~Cosma, R.~T. Ionescu, and M.~Popescu, ``Black-{Box} ripper: copying black-box models using generative evolutionary algorithms,'' in \emph{Proceedings of the 34th {International} {Conference} on {Neural} {Information} {Processing} {Systems}}, ser. {NIPS} '20.\hskip 1em plus 0.5em minus 0.4em\relax Red Hook, NY, USA: Curran Associates Inc., 2020, event-place: Vancouver, BC, Canada.

\bibitem{yu_cloudleak_2020}
\BIBentryALTinterwordspacing
H.~Yu, K.~Yang, T.~Zhang, Y.-Y. Tsai, T.-Y. Ho, and Y.~Jin, ``\BIBforeignlanguage{en}{{CloudLeak}: {Large}-{Scale} {Deep} {Learning} {Models} {Stealing} {Through} {Adversarial} {Examples}},'' in \emph{\BIBforeignlanguage{en}{Proceedings 2020 {Network} and {Distributed} {System} {Security} {Symposium}}}.\hskip 1em plus 0.5em minus 0.4em\relax San Diego, CA: Internet Society, 2020. [Online]. Available: \url{https://www.ndss-symposium.org/wp-content/uploads/2020/02/24178.pdf}
\BIBentrySTDinterwordspacing

\bibitem{gong_inversenet_2021}
\BIBentryALTinterwordspacing
X.~Gong, Y.~Chen, W.~Yang, G.~Mei, and Q.~Wang, ``\BIBforeignlanguage{en}{{InverseNet}: {Augmenting} {Model} {Extraction} {Attacks} with {Training} {Data} {Inversion}},'' in \emph{\BIBforeignlanguage{en}{Proceedings of the {Thirtieth} {International} {Joint} {Conference} on {Artificial} {Intelligence}}}.\hskip 1em plus 0.5em minus 0.4em\relax Montreal, Canada: International Joint Conferences on Artificial Intelligence Organization, Aug. 2021, pp. 2439--2447. [Online]. Available: \url{https://www.ijcai.org/proceedings/2021/336}
\BIBentrySTDinterwordspacing

\bibitem{truong_data-free_2021}
J.-B. Truong, P.~Maini, R.~J. Walls, and N.~Papernot, ``Data-{Free} {Model} {Extraction},'' in \emph{Proceedings of the {IEEE}/{CVF} {Conference} on {Computer} {Vision} and {Pattern} {Recognition} ({CVPR})}, Jun. 2021, pp. 4771--4780.

\bibitem{miura_megex_2024}
\BIBentryALTinterwordspacing
T.~Miura, T.~Shibahara, and N.~Yanai, ``\BIBforeignlanguage{en}{{MEGEX}: {Data}-{Free} {Model} {Extraction} {Attack} {Against} {Gradient}-{Based} {Explainable} {AI}},'' in \emph{\BIBforeignlanguage{en}{Proceedings of the 2nd {ACM} {Workshop} on {Secure} and {Trustworthy} {Deep} {Learning} {Systems}}}.\hskip 1em plus 0.5em minus 0.4em\relax Singapore Singapore: ACM, Jul. 2024, pp. 56--66. [Online]. Available: \url{https://dl.acm.org/doi/10.1145/3665451.3665533}
\BIBentrySTDinterwordspacing

\bibitem{sanyal_towards_2022}
\BIBentryALTinterwordspacing
S.~Sanyal, S.~Addepalli, and R.~V. Babu, ``Towards {Data}-{Free} {Model} {Stealing} in a {Hard} {Label} {Setting},'' in \emph{2022 {IEEE}/{CVF} {Conference} on {Computer} {Vision} and {Pattern} {Recognition} ({CVPR})}.\hskip 1em plus 0.5em minus 0.4em\relax New Orleans, LA, USA: IEEE, Jun. 2022, pp. 15\,263--15\,272. [Online]. Available: \url{https://ieeexplore.ieee.org/document/9880326/}
\BIBentrySTDinterwordspacing

\bibitem{zhang_thief_2021}
\BIBentryALTinterwordspacing
X.~Zhang, C.~Fang, and J.~Shi, ``Thief, {Beware} of {What} {Get} {You} {There}: {Towards} {Understanding} {Model} {Extraction} {Attack},'' 2021, version Number: 1. [Online]. Available: \url{https://arxiv.org/abs/2104.05921}
\BIBentrySTDinterwordspacing

\bibitem{wang_enhance_2022}
\BIBentryALTinterwordspacing
Y.~Wang and X.~Lin, ``Enhance {Model} {Stealing} {Attack} via {Label} {Refining},'' in \emph{2022 7th {International} {Conference} on {Intelligent} {Computing} and {Signal} {Processing} ({ICSP})}.\hskip 1em plus 0.5em minus 0.4em\relax Xi'an, China: IEEE, Apr. 2022, pp. 1040--1043. [Online]. Available: \url{https://ieeexplore.ieee.org/document/9778562/}
\BIBentrySTDinterwordspacing

\bibitem{wang_black-box_2022}
\BIBentryALTinterwordspacing
Y.~Wang, J.~Li, H.~Liu, Y.~Wang, Y.~Wu, F.~Huang, and R.~Ji, ``Black-{Box} {Dissector}: {Towards} {Erasing}-{Based} {Hard}-{Label} {Model} {Stealing} {Attack},'' in \emph{Computer {Vision} – {ECCV} 2022: 17th {European} {Conference}, {Tel} {Aviv}, {Israel}, {October} 23–27, 2022, {Proceedings}, {Part} {V}}.\hskip 1em plus 0.5em minus 0.4em\relax Berlin, Heidelberg: Springer-Verlag, 2022, pp. 192--208, event-place: Tel Aviv, Israel. [Online]. Available: \url{https://doi.org/10.1007/978-3-031-20065-6_12}
\BIBentrySTDinterwordspacing

\bibitem{yan_towards_2022}
\BIBentryALTinterwordspacing
A.~Yan, R.~Hou, X.~Liu, H.~Yan, T.~Huang, and X.~Wang, ``\BIBforeignlanguage{en}{Towards explainable model extraction attacks},'' \emph{\BIBforeignlanguage{en}{International Journal of Intelligent Systems}}, vol.~37, no.~11, pp. 9936--9956, Nov. 2022. [Online]. Available: \url{https://onlinelibrary.wiley.com/doi/10.1002/int.23022}
\BIBentrySTDinterwordspacing

\bibitem{xie_game_2022}
Y.~Xie, M.~Huang, X.~Zhang, C.~Dong, W.~Susilo, and X.~Chen, ``{GAME}: {Generative}-{Based} {Adaptive} {Model} {Extraction} {Attack},'' in \emph{Computer {Security} – {ESORICS} 2022}, V.~Atluri, R.~Di~Pietro, C.~D. Jensen, and W.~Meng, Eds.\hskip 1em plus 0.5em minus 0.4em\relax Cham: Springer International Publishing, 2022, pp. 570--588.

\bibitem{chen_d-dae_2023}
\BIBentryALTinterwordspacing
Y.~Chen, R.~Guan, X.~Gong, J.~Dong, and M.~Xue, ``D-{DAE}: {Defense}-{Penetrating} {Model} {Extraction} {Attacks},'' in \emph{2023 {IEEE} {Symposium} on {Security} and {Privacy} ({SP})}.\hskip 1em plus 0.5em minus 0.4em\relax San Francisco, CA, USA: IEEE, May 2023, pp. 382--399. [Online]. Available: \url{https://ieeexplore.ieee.org/document/10179406/}
\BIBentrySTDinterwordspacing

\bibitem{he_drmi_2021}
\BIBentryALTinterwordspacing
Y.~He, G.~Meng, K.~Chen, X.~Hu, and J.~He, ``{DRMI}: {A} {Dataset} {Reduction} {Technology} based on {Mutual} {Information} for {Black}-box {Attacks},'' in \emph{30th {USENIX} {Security} {Symposium} ({USENIX} {Security} 21)}.\hskip 1em plus 0.5em minus 0.4em\relax USENIX Association, Aug. 2021, pp. 1901--1918. [Online]. Available: \url{https://www.usenix.org/conference/usenixsecurity21/presentation/he-yingzhe}
\BIBentrySTDinterwordspacing

\bibitem{liu_ml-doctor_2022}
\BIBentryALTinterwordspacing
Y.~Liu, R.~Wen, X.~He, A.~Salem, Z.~Zhang, M.~Backes, E.~D. Cristofaro, M.~Fritz, and Y.~Zhang, ``{ML}-{Doctor}: {Holistic} {Risk} {Assessment} of {Inference} {Attacks} {Against} {Machine} {Learning} {Models},'' in \emph{31st {USENIX} {Security} {Symposium} ({USENIX} {Security} 22)}.\hskip 1em plus 0.5em minus 0.4em\relax Boston, MA: USENIX Association, Aug. 2022, pp. 4525--4542. [Online]. Available: \url{https://www.usenix.org/conference/usenixsecurity22/presentation/liu-yugeng}
\BIBentrySTDinterwordspacing

\bibitem{rosenthal_disguide_2023}
\BIBentryALTinterwordspacing
J.~Rosenthal, E.~Enouen, H.~V. Pham, and L.~Tan, ``{DisGUIDE}: {Disagreement}-{Guided} {Data}-{Free} {Model} {Extraction},'' \emph{Proceedings of the AAAI Conference on Artificial Intelligence}, vol.~37, no.~8, pp. 9614--9622, Jun. 2023. [Online]. Available: \url{https://ojs.aaai.org/index.php/AAAI/article/view/26150}
\BIBentrySTDinterwordspacing

\bibitem{yan_holistic_2023}
\BIBentryALTinterwordspacing
A.~Yan, H.~Yan, L.~Hu, X.~Liu, and T.~Huang, ``Holistic {Implicit} {Factor} {Evaluation} of {Model} {Extraction} {Attacks},'' \emph{IEEE Transactions on Dependable and Secure Computing}, vol.~20, no.~6, pp. 4678--4689, Nov. 2023. [Online]. Available: \url{https://ieeexplore.ieee.org/document/9999271/}
\BIBentrySTDinterwordspacing

\bibitem{zhang_towards_2022}
\BIBentryALTinterwordspacing
J.~Zhang, B.~Li, J.~Xu, S.~Wu, S.~Ding, L.~Zhang, and C.~Wu, ``Towards {Efficient} {Data} {Free} {Blackbox} {Adversarial} {Attack},'' in \emph{2022 {IEEE}/{CVF} {Conference} on {Computer} {Vision} and {Pattern} {Recognition} ({CVPR})}.\hskip 1em plus 0.5em minus 0.4em\relax New Orleans, LA, USA: IEEE, Jun. 2022, pp. 15\,094--15\,104. [Online]. Available: \url{https://ieeexplore.ieee.org/document/9880352/}
\BIBentrySTDinterwordspacing

\bibitem{yan_explanation-based_2023}
\BIBentryALTinterwordspacing
A.~Yan, R.~Hou, H.~Yan, and X.~Liu, ``\BIBforeignlanguage{en}{Explanation-based data-free model extraction attacks},'' \emph{\BIBforeignlanguage{en}{World Wide Web}}, vol.~26, no.~5, pp. 3081--3092, Sep. 2023. [Online]. Available: \url{https://link.springer.com/10.1007/s11280-023-01150-6}
\BIBentrySTDinterwordspacing

\bibitem{yan_explanation_2023}
\BIBentryALTinterwordspacing
A.~Yan, T.~Huang, L.~Ke, X.~Liu, Q.~Chen, and C.~Dong, ``\BIBforeignlanguage{en}{Explanation leaks: {Explanation}-guided model extraction attacks},'' \emph{\BIBforeignlanguage{en}{Information Sciences}}, vol. 632, pp. 269--284, Jun. 2023. [Online]. Available: \url{https://linkinghub.elsevier.com/retrieve/pii/S002002552300316X}
\BIBentrySTDinterwordspacing

\bibitem{yang_efficient_2023}
\BIBentryALTinterwordspacing
P.~Yang, Q.~Wu, and X.~Zhang, ``Efficient {Model} {Extraction} by {Data} {Set} {Stealing}, {Balancing}, and {Filtering},'' \emph{IEEE Internet of Things Journal}, vol.~10, no.~24, pp. 22\,717--22\,725, Dec. 2023. [Online]. Available: \url{https://ieeexplore.ieee.org/document/10214537/}
\BIBentrySTDinterwordspacing

\bibitem{lin_quda_2023}
\BIBentryALTinterwordspacing
Z.~Lin, K.~Xu, C.~Fang, H.~Zheng, A.~Ahmed~Jaheezuddin, and J.~Shi, ``\BIBforeignlanguage{en}{{QUDA}: {Query}-{Limited} {Data}-{Free} {Model} {Extraction}},'' in \emph{\BIBforeignlanguage{en}{Proceedings of the {ACM} {Asia} {Conference} on {Computer} and {Communications} {Security}}}.\hskip 1em plus 0.5em minus 0.4em\relax Melbourne VIC Australia: ACM, Jul. 2023, pp. 913--924. [Online]. Available: \url{https://dl.acm.org/doi/10.1145/3579856.3590336}
\BIBentrySTDinterwordspacing

\bibitem{liu_shrewdattack_2023}
\BIBentryALTinterwordspacing
Y.~Liu, J.~Luo, Y.~Yang, X.~Wang, M.~Gheisari, and F.~Luo, ``\BIBforeignlanguage{en}{{ShrewdAttack}: {Low} {Cost} {High} {Accuracy} {Model} {Extraction}},'' \emph{\BIBforeignlanguage{en}{Entropy}}, vol.~25, no.~2, p. 282, Feb. 2023. [Online]. Available: \url{https://www.mdpi.com/1099-4300/25/2/282}
\BIBentrySTDinterwordspacing

\bibitem{pape_limitations_2023}
\BIBentryALTinterwordspacing
D.~Pape, S.~Däubener, T.~Eisenhofer, A.~E. Cinà, and L.~Schönherr, ``On the {Limitations} of {Model} {Stealing} with {Uncertainty} {Quantification} {Models},'' in \emph{The {Second} {Workshop} on {New} {Frontiers} in {Adversarial} {Machine} {Learning}}, 2023. [Online]. Available: \url{https://openreview.net/forum?id=ONRFHoUzNk}
\BIBentrySTDinterwordspacing

\bibitem{liu_efficient_2024}
\BIBentryALTinterwordspacing
Y.~Liu, R.~Wen, M.~Backes, and Y.~Zhang, ``Efficient {Data}-{Free} {Model} {Stealing} with {Label} {Diversity},'' 2024, version Number: 1. [Online]. Available: \url{https://arxiv.org/abs/2404.00108}
\BIBentrySTDinterwordspacing

\bibitem{khaled_careful_2022}
\BIBentryALTinterwordspacing
K.~Khaled, G.~Nicolescu, and F.~G. De~Magalhaes, ``Careful {What} {You} {Wish} {For}: on the {Extraction} of {Adversarially} {Trained} {Models},'' in \emph{2022 19th {Annual} {International} {Conference} on {Privacy}, {Security} \& {Trust} ({PST})}.\hskip 1em plus 0.5em minus 0.4em\relax Fredericton, NB, Canada: IEEE, Aug. 2022, pp. 1--10. [Online]. Available: \url{https://ieeexplore.ieee.org/document/9851981/}
\BIBentrySTDinterwordspacing

\bibitem{zhao_extracting_2023}
\BIBentryALTinterwordspacing
S.~Zhao, K.~Chen, M.~Hao, J.~Zhang, G.~Xu, H.~Li, and T.~Zhang, ``Extracting {Cloud}-based {Model} with {Prior} {Knowledge},'' 2023, version Number: 4. [Online]. Available: \url{https://arxiv.org/abs/2306.04192}
\BIBentrySTDinterwordspacing

\bibitem{karmakar_marich_2023}
\BIBentryALTinterwordspacing
P.~Karmakar and D.~Basu, ``Marich: {A} {Query}-efficient {Distributionally} {Equivalent} {Model} {Extraction} {Attack},'' in \emph{Advances in {Neural} {Information} {Processing} {Systems}}, A.~Oh, T.~Naumann, A.~Globerson, K.~Saenko, M.~Hardt, and S.~Levine, Eds., vol.~36.\hskip 1em plus 0.5em minus 0.4em\relax Curran Associates, Inc., 2023, pp. 72\,412--72\,445. [Online]. Available: \url{https://proceedings.neurips.cc/paper_files/paper/2023/file/e5440ffceaf4831b5f98652b8a27ffde-Paper-Conference.pdf}
\BIBentrySTDinterwordspacing

\bibitem{jindal_army_2024}
\BIBentryALTinterwordspacing
A.~Jindal, V.~Goyal, S.~Anand, and C.~Arora, ``Army of {Thieves}: {Enhancing} {Black}-{Box} {Model} {Extraction} via {Ensemble} based sample selection,'' in \emph{2024 {IEEE}/{CVF} {Winter} {Conference} on {Applications} of {Computer} {Vision} ({WACV})}.\hskip 1em plus 0.5em minus 0.4em\relax Waikoloa, HI, USA: IEEE, Jan. 2024, pp. 3811--3820. [Online]. Available: \url{https://ieeexplore.ieee.org/document/10484430/}
\BIBentrySTDinterwordspacing

\bibitem{beetham_dual_2023}
\BIBentryALTinterwordspacing
J.~Beetham, N.~Kardan, A.~S. Mian, and M.~Shah, ``Dual {Student} {Networks} for {Data}-{Free} {Model} {Stealing},'' in \emph{The {Eleventh} {International} {Conference} on {Learning} {Representations}}, 2023. [Online]. Available: \url{https://openreview.net/forum?id=VE1s3e5xriA}
\BIBentrySTDinterwordspacing

\bibitem{hong_exploring_2023}
\BIBentryALTinterwordspacing
C.~Hong, J.~Huang, R.~Birke, and L.~Y. Chen, ``Exploring and {Exploiting} {Data}-{Free} {Model} {Stealing},'' in \emph{Machine {Learning} and {Knowledge} {Discovery} in {Databases}: {Research} {Track}: {European} {Conference}, {ECML} {PKDD} 2023, {Turin}, {Italy}, {September} 18–22, 2023, {Proceedings}, {Part} {V}}.\hskip 1em plus 0.5em minus 0.4em\relax Berlin, Heidelberg: Springer-Verlag, 2023, pp. 20--35, event-place: Turin, Italy. [Online]. Available: \url{https://doi.org/10.1007/978-3-031-43424-2_2}
\BIBentrySTDinterwordspacing

\bibitem{yang_swifttheft_2024}
\BIBentryALTinterwordspacing
W.~Yang, X.~Gong, Y.~Chen, Q.~Wang, and J.~Dong, ``{SwiftTheft}: {A} {Time}-{Efficient} {Model} {Extraction} {Attack} {Framework} {Against} {Cloud}-{Based} {Deep} {Neural} {Networks},'' \emph{Chinese Journal of Electronics}, vol.~33, no.~1, pp. 90--100, Jan. 2024. [Online]. Available: \url{https://ieeexplore.ieee.org/document/10410588/}
\BIBentrySTDinterwordspacing

\bibitem{fredrikson_model_2015}
\BIBentryALTinterwordspacing
M.~Fredrikson, S.~Jha, and T.~Ristenpart, ``\BIBforeignlanguage{en}{Model {Inversion} {Attacks} that {Exploit} {Confidence} {Information} and {Basic} {Countermeasures}},'' in \emph{\BIBforeignlanguage{en}{Proceedings of the 22nd {ACM} {SIGSAC} {Conference} on {Computer} and {Communications} {Security}}}.\hskip 1em plus 0.5em minus 0.4em\relax Denver Colorado USA: ACM, Oct. 2015, pp. 1322--1333. [Online]. Available: \url{https://dl.acm.org/doi/10.1145/2810103.2813677}
\BIBentrySTDinterwordspacing

\bibitem{chandrasekaran_exploring_2020}
\BIBentryALTinterwordspacing
V.~Chandrasekaran, K.~Chaudhuri, I.~Giacomelli, S.~Jha, and S.~Yan, ``Exploring {Connections} {Between} {Active} {Learning} and {Model} {Extraction},'' in \emph{29th {USENIX} {Security} {Symposium} ({USENIX} {Security} 20)}.\hskip 1em plus 0.5em minus 0.4em\relax USENIX Association, Aug. 2020, pp. 1309--1326. [Online]. Available: \url{https://www.usenix.org/conference/usenixsecurity20/presentation/chandrasekaran}
\BIBentrySTDinterwordspacing

\bibitem{krizhevsky_learning_2009}
A.~Krizhevsky and G.~Hinton, ``\BIBforeignlanguage{en}{Learning {Multiple} {Layers} of {Features} from {Tiny} {Images}},'' University of Toronto, Tech. Rep., 2009.

\bibitem{hasanpour_lets_2016}
\BIBentryALTinterwordspacing
S.~H. Hasanpour, M.~Rouhani, M.~Fayyaz, and M.~Sabokrou, ``Lets keep it simple, {Using} simple architectures to outperform deeper and more complex architectures,'' 2016, version Number: 8. [Online]. Available: \url{https://arxiv.org/abs/1608.06037}
\BIBentrySTDinterwordspacing

\bibitem{grana_perturbing_2020}
\BIBentryALTinterwordspacing
J.~Grana, ``Perturbing {Inputs} to {Prevent} {Model} {Stealing},'' in \emph{2020 {IEEE} {Conference} on {Communications} and {Network} {Security} ({CNS})}.\hskip 1em plus 0.5em minus 0.4em\relax Avignon, France: IEEE, Jun. 2020, pp. 1--9. [Online]. Available: \url{https://ieeexplore.ieee.org/document/9162336/}
\BIBentrySTDinterwordspacing

\bibitem{cao_mitigating_2017}
\BIBentryALTinterwordspacing
X.~Cao and N.~Z. Gong, ``\BIBforeignlanguage{en}{Mitigating {Evasion} {Attacks} to {Deep} {Neural} {Networks} via {Region}-based {Classification}},'' in \emph{\BIBforeignlanguage{en}{Proceedings of the 33rd {Annual} {Computer} {Security} {Applications} {Conference}}}.\hskip 1em plus 0.5em minus 0.4em\relax Orlando FL USA: ACM, Dec. 2017, pp. 278--287. [Online]. Available: \url{https://dl.acm.org/doi/10.1145/3134600.3134606}
\BIBentrySTDinterwordspacing

\bibitem{orekondy_prediction_2020}
\BIBentryALTinterwordspacing
T.~Orekondy, B.~Schiele, and M.~Fritz, ``Prediction {Poisoning}: {Towards} {Defenses} {Against} {DNN} {Model} {Stealing} {Attacks},'' in \emph{International {Conference} on {Learning} {Representations}}, ser. {ICLR}, Virtual Event, Apr. 2020. [Online]. Available: \url{https://iclr.cc/virtual_2020/poster_SyevYxHtDB.html}
\BIBentrySTDinterwordspacing

\bibitem{chen_das-ast_2021}
J.~Chen, C.~Wu, S.~Shen, X.~Zhang, and J.~Chen, ``{DAS}-{AST}: {Defending} {Against} {Model} {Stealing} {Attacks} {Based} on {Adaptive} {Softmax} {Transformation},'' in \emph{Information {Security} and {Cryptology}}, Y.~Wu and M.~Yung, Eds.\hskip 1em plus 0.5em minus 0.4em\relax Cham: Springer International Publishing, 2021, pp. 21--36.

\bibitem{mazeika_how_2022}
\BIBentryALTinterwordspacing
M.~Mazeika, B.~Li, and D.~Forsyth, ``How to {Steer} {Your} {Adversary}: {Targeted} and {Efficient} {Model} {Stealing} {Defenses} with {Gradient} {Redirection},'' in \emph{Proceedings of the 39th {International} {Conference} on {Machine} {Learning}}, ser. Proceedings of {Machine} {Learning} {Research}, K.~Chaudhuri, S.~Jegelka, L.~Song, C.~Szepesvari, G.~Niu, and S.~Sabato, Eds., vol. 162.\hskip 1em plus 0.5em minus 0.4em\relax PMLR, Jul. 2022, pp. 15\,241--15\,254. [Online]. Available: \url{https://proceedings.mlr.press/v162/mazeika22a.html}
\BIBentrySTDinterwordspacing

\bibitem{ducoffe_adversarial_2018}
\BIBentryALTinterwordspacing
M.~Ducoffe and F.~Precioso, ``Adversarial {Active} {Learning} for {Deep} {Networks}: a {Margin} {Based} {Approach},'' 2018, version Number: 1. [Online]. Available: \url{https://arxiv.org/abs/1802.09841}
\BIBentrySTDinterwordspacing

\bibitem{sener_active_2018}
\BIBentryALTinterwordspacing
O.~Sener and S.~Savarese, ``Active {Learning} for {Convolutional} {Neural} {Networks}: {A} {Core}-{Set} {Approach},'' in \emph{International {Conference} on {Learning} {Representations}}, 2018. [Online]. Available: \url{https://openreview.net/forum?id=H1aIuk-RW}
\BIBentrySTDinterwordspacing

\bibitem{moosavi-dezfooli_deepfool_2016}
\BIBentryALTinterwordspacing
S.-M. Moosavi-Dezfooli, A.~Fawzi, and P.~Frossard, ``{DeepFool}: {A} {Simple} and {Accurate} {Method} to {Fool} {Deep} {Neural} {Networks},'' in \emph{2016 {IEEE} {Conference} on {Computer} {Vision} and {Pattern} {Recognition} ({CVPR})}.\hskip 1em plus 0.5em minus 0.4em\relax Las Vegas, NV, USA: IEEE, Jun. 2016, pp. 2574--2582. [Online]. Available: \url{http://ieeexplore.ieee.org/document/7780651/}
\BIBentrySTDinterwordspacing

\end{thebibliography}

\clearpage

\appendix
\subsection{Attack Setup Overview}
\label{appendix:attack_setup}
\begin{figure*}[hb!]
\centering
\includesvg[width=0.8\textwidth]{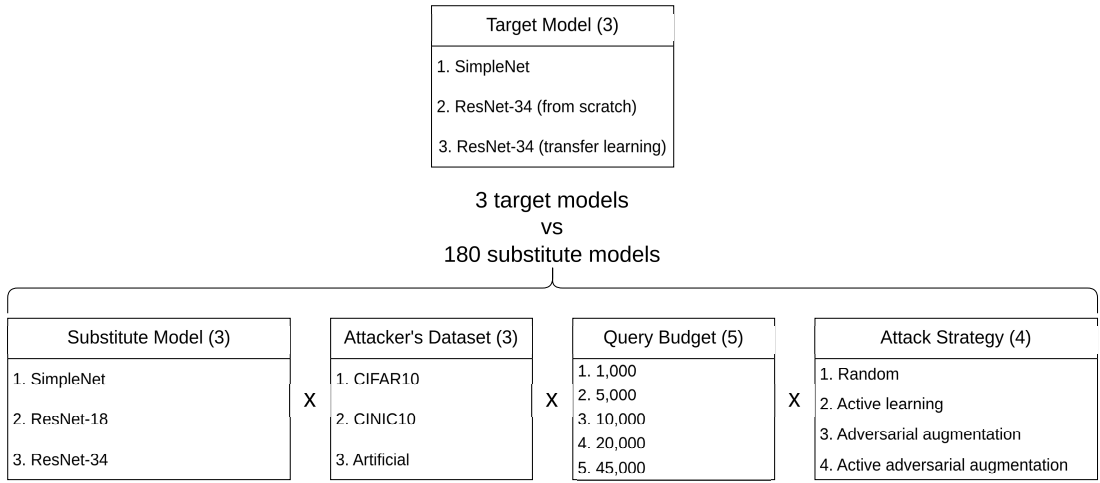}
\caption{Overview of attack setups examined in this work.} 
\label{fig:attack_setup}
\end{figure*}

\Cref{fig:attack_setup} illustrates an overview of all conducted attacks. Overall, we trained 180 substitute models against each target model.


\subsection{Query Optimisation Methods}
\label{appendix:query_optimisation_methods}
\subsubsection{Active Learning}

In active learning, there is a model that has to be trained on data labelled by an oracle. For model stealing attacks, a substitute model $\hat{f}$ corresponds to that trainable model, and the target model $f$ corresponds to the oracle. Further, both seed (labelled data) and pool (unlabelled) data belong to the attacker's data. Hence, at each active learning round, the target model labels a certain amount of the attacker's data.

The active learning optimisation strategy used in this work is a slightly modified version of an approach from previous work by Pal et al. \cite{pal_activethief_2020}. The authors compared several active learning strategies for image and text classification. We selected the strategy with the best performance rate on image classification tasks. It combines two active learning algorithms, namely the Deepfool Active Learning (DFAL) \cite{ducoffe_adversarial_2018}, and the $\kappa$-center algorithm \cite{sener_active_2018}. 
Pal et al. \cite{pal_activethief_2020} combined the algorithms in the following way. In each active learning round, they applied DFAL to select $q$ samples, where $q$ corresponds to the total query budget. Then they applied $\kappa$-center to select $k$ most suitable samples out of $q$. In this work, in order to decrease the computation time and make the attack more efficient, we made the following changes: (i) reducing the number of samples selected by DFAL to $2k$, and (ii) splitting the pool into several sub-pools, so that at each round data is selected from a single sub-pool. The latter modification is applied to all optimisation techniques.

We summarise the stealing process with active learning in \Cref{alg:al_attack}. As for any model stealing attack, we need target and substitute models, the attacker's data (pool), and the query budget. Additionally, unlike in the previous work, we have two hyperparameters specific to attacks with query optimisation: seed size and the number of rounds, which are tuned on a validation dataset of the attacker's dataset.  We assume that the seed is randomly selected from the pool, and the substitute model is trained on it. The value of $k$ (number of samples to select per round) is calculated based on the query budget, seed size, and the number of rounds. At each round, we consequently apply DFAL and $\kappa$-center, add selected samples to the seed, and, as suggested in previous work, train the substitute model from scratch on the augmented dataset. 

\begin{algorithm}[th]
    \caption{Active learning attack}\label{alg:al_attack}
    \SetKwFunction{train}{Train}
    \SetKwFunction{dfal}{DFAL}
    \SetKwFunction{kcenter}{$\kappa$-center}
    \KwIn{target model $f$, substitute model $\hat{f}$, data pool $D$, query budget $q$, seed size $s$, number of rounds $r$}
    \KwOut{$\hat{f}$}
    $S_0 \gets$ select randomly $s$ samples from $D$ \;
    $y_0 \gets f(S_0)$ \;
    $\hat{f} \gets \train(\hat{f}, S_0, y_0)$ \;
    $k \gets \dfrac{q - |S_0|}{r}$ \;
    \For{$i=1$ to $r$}{
        $x_1 \ldots x_{2k} \gets \dfal(\hat{f}, D, 2k)$ \;
        $x_1^{\prime} \ldots x_{k}^{\prime} \gets \kcenter(\hat{f}, S_{i-1}, \{x_1 \ldots x_{2k}\}, k)$ \;
        $D \gets D \setminus \{x_1^{\prime}, \ldots, x_k^{\prime}\}$ \;
        $S_i \gets S_{i-1} \cup \{x_1^{\prime}, \ldots, x_k^{\prime}\}$ \;
        $y_i \gets y_{i-1} \cup \{f(x_1^{\prime}), \ldots, f(x_k^{\prime})\}$ \;
        $\hat{f} \gets \train(\hat{f}, S_{i}, y_i)$ \;
        }
    \end{algorithm}    

\subsubsection{Adversarial Augmentation}
The second query optimisation technique used in this work is adversarial augmentation. Instead of selecting the most promising samples as active learning, adversarial augmentation modifies samples to make them more information-revealing. Since adversarial examples lay close to the decision boundary, it is assumed that they should help better approximate the target model's decision boundary. Similarly to active learning, we picked an adversarial example crafting strategy based on previous work. Pengcheng et al. \cite{pengcheng_query-efficient_2018} compared different adversarial crafting techniques for model stealing targeting image classifiers. As Deepfool's \cite{moosavi-dezfooli_deepfool_2016} performance was the most promising, we selected it for our adversarial augmentation attack. 

We summarise our implemented attack in \Cref{alg:adv_attack}. The input parameters are the same as for the active learning attack. At each augmentation round, we randomly select samples from the pool, which are then augmented with their Deepfool adversarial examples. Both clean and adversarial samples are then labelled by the target model and added to the seed. The substitute model is trained from scratch after each round.

\begin{algorithm}[th]
    \caption{Adversarial augmentation attack}\label{alg:adv_attack}
    \SetKwFunction{train}{Train}
    \SetKwFunction{deepfool}{Deepfool}
    \KwIn{target model $f$, substitute model $\hat{f}$, data pool $D$, query budget $q$, seed size $s$, number of rounds $r$}
    \KwOut{$\hat{f}$}
    $S_0 \gets$ select randomly $s$ samples from $D$ \;
    $y_0 \gets f(S_0)$ \;
    $\hat{f} \gets \train(\hat{f}, S_0, y_0)$ \;
    $k \gets \dfrac{q - |S_0|}{r}$ \;
    \For{$i=1$ to $r$}{
        $x_1^{\prime} \ldots x_{\frac{k}{2}}^{\prime} \gets$ select randomly $\frac{k}{2}$ samples from $D$ \;
        $x_{\frac{k}{2}+1}^{\prime} \ldots x_{k}^{\prime} \gets \deepfool(\hat{f}, x_1^{\prime}), \ldots, \deepfool(\hat{f}, x_{\frac{k}{2}}^{\prime})$ \;
        $D \gets D \setminus \{x_1^{\prime}, \ldots, x_k^{\prime}\}$ \;
        $S_i \gets S_{i-1} \cup \{x_1^{\prime}, \ldots, x_k^{\prime}\}$ \;
        $y_i \gets y_{i-1} \cup \{f(x_1^{\prime}), \ldots, f(x_k^{\prime})\}$ \;
        $\hat{f} \gets \train(\hat{f}, S_{i}, y_i)$ \;
        }
    \end{algorithm}    

\subsubsection{Active Adversarial Augmentation}

The combined attack is shown in \Cref{alg:aaa_attack}. The algorithm repeats the behaviour of the adversarial augmentation attack (\Cref{alg:adv_attack}) with one difference: line 6 in \Cref{alg:adv_attack} is replaced with lines 6 and 7 in \Cref{alg:aaa_attack}. Instead of randomly selecting $\frac{k}{2}$ samples out of the pool, they are now selected using active learning algorithms.

\begin{algorithm}[th]
    \caption{Active adversarial augmentation attack}\label{alg:aaa_attack}
    \SetKwFunction{train}{Train}
    \SetKwFunction{dfal}{DFAL}
    \SetKwFunction{kcenter}{$\kappa$-center}
    \SetKwFunction{deepfool}{Deepfool}
    \KwIn{target model $f$, substitute model $\hat{f}$, data pool $D$, query budget $q$, seed size $s$, number of rounds $r$}
    \KwOut{$\hat{f}$}
    $S_0 \gets$ select randomly $s$ samples from $D$ \;
    $y_0 \gets f(S_0)$ \;
    $\hat{f} \gets \train(\hat{f}, S_0, y_0)$ \;
    $k \gets \dfrac{q - |S_0|}{r}$ \;
    \For{$i=1$ to $r$}{
        $x_1 \ldots x_{k} \gets \dfal(\hat{f}, D, k)$ \;
        $x_1^{\prime} \ldots x_{\frac{k}{2}}^{\prime} \gets \kcenter(\hat{f}, S_{i-1}, \{x_1 \ldots x_{k}\}, \frac{k}{2})$ \;
        $x_{\frac{k}{2}+1}^{\prime} \ldots x_{k}^{\prime} \gets \deepfool(\hat{f}, x_1^{\prime}), \ldots, \deepfool(\hat{f}, x_{\frac{k}{2}}^{\prime})$ \;
        $D \gets D \setminus \{x_1^{\prime}, \ldots, x_k^{\prime}\}$ \;
        $S_i \gets S_{i-1} \cup \{x_1^{\prime}, \ldots, x_k^{\prime}\}$ \;
        $y_i \gets y_{i-1} \cup \{f(x_1^{\prime}), \ldots, f(x_k^{\prime})\}$ \;
        $\hat{f} \gets \train(\hat{f}, S_{i}, y_i)$ \;
        }
    \end{algorithm}    


\subsection{Value Estimation for the State-of-the-Art Comparison}
\label{appendix:value_estimation_for_sota}
As mentioned in \Cref{sec:comparison_sota}, some of the values in \Cref{tab:attacks_vs_sota_below_40} and \Cref{tab:attacks_vs_sota_45+} were estimated, as the exact values were not provided by the authors. Below, we explain how each value was configured and which information was used for the estimation. 
\begin{itemize}
    \item The target and substitute accuracy for \cite{milli_model_2019}, substitute accuracy for \cite{yan_towards_2022, yan_explanation-based_2023}, as well as substitute accuracy and fidelity for \cite{khaled_careful_2022} were estimated from plots, as no actual scores were reported.
    \item The number of queries used by \cite{pape_limitations_2023}
    is estimated as 25,000 queries as the authors mention using half of the (CIFAR-10) training set for an attack, but explicitly the number is not confirmed. 
    \item The number of queries used by \cite{barbalau_black-box_2020} was not reported. However, the authors used CIFAR-100, which contains 50,000 samples, as a \textit{starting point} for their evolutionary algorithm that creates new query images. They claimed that optimising the query budget was not a priority, so we can assume that the number of queries could be significantly larger than 50,000. However, as that is only a speculation, we conservatively estimate that they used "more than 50,000" samples. 
    \item The authors of \cite{mosafi_stealing_2019} also do not provide information about the total number of queries. However, they have an iterative algorithm that generates 1,000,000 samples at each round. Assuming that there should be at least one round, we estimated the number of queries as "larger than 1,000,000".
    \item The query budget was also not given by \cite{yuan_es_2022}. However, the authors estimated the price of their attack on Amazon Web Services\footnote{\url{https://aws.amazon.com/rekognition/pricing/}} to be \$360,000. Pricing from 2023 suggests that the price per query for the first million queries is usually around \$0.001, getting cheaper for the subsequent millions. That means that a million queries cost \$1000, and with this price, the authors could have made more than 300 million queries. However, since the prices could differ back then, we lowered our estimate to 1 million. 
    \item Some of the papers did not report the architecture of the target model or used a custom architecture without reporting the number of trainable parameters. In these cases, we could not estimate the number of parameters of the target model and, consequently, could not report the efficiency score of an attack. In cases when we estimated the number of queries, we also marked the efficiency score as estimated. In one paper \cite{pengcheng_query-efficient_2018}, the authors only mention that they use a ResNet model, referring to the paper where ResNet models were introduced. As this original paper mainly focuses on ResNet-34 architecture, we assumed that ResNet-34 was used as a target architecture by \cite{pengcheng_query-efficient_2018} and calculated efficiency scores based on this assumption.
    
\end{itemize}

\subsection{Additional Results}
\label{appendix:additional_results}

\Cref{tab:attacks_substitute_arch_cinic} and \Cref{tab:attacks_substitute_arch_artificial} show the performance of substitute models with different architectures trained on PD (CINIC-10) and artificial data.

\begin{table*}[t]
\centering
\caption{Performance of substitute models with different architectures trained on PD (CINIC-10) data.}
\begin{tabular}{ccccccccccc}
\toprule
                            & \textbf{Target $\rightarrow$}                                                           & \multicolumn{3}{c}{\textbf{SimpleNet}}                                                                                                              & \multicolumn{3}{c}{\begin{tabular}[c]{@{}c@{}}\textbf{ResNet-34} \\ \textbf{(from scratch)}\end{tabular}}                                                  & \multicolumn{3}{c}{\begin{tabular}[c]{@{}c@{}}\textbf{ResNet-34} \\ \textbf{(transfer learning)}\end{tabular}}                                              \\ \cmidrule(lr){3-5} \cmidrule(lr){6-8} \cmidrule(lr){9-11}
\textbf{Substitute $\downarrow$}     &  \textit{QB}  & \textit{Joint Acc} & \textit{Acc}                                      & \textit{Fid}                                      & \textit{Joint Acc} & \textit{Acc}                                      & \textit{Fid}                                     & \textit{Joint Acc} & \textit{Acc}                                      & \textit{Fid}                                      \\ \midrule
                            & 1k                                                                             & \cellcolor[HTML]{EFEFEF}36.66\%                      & \cellcolor[HTML]{EFEFEF}38.62\%          & \cellcolor[HTML]{EFEFEF}38.88\%          & 39.83\%                                             & 41.28\%                                  & 41.62\%                                  & \cellcolor[HTML]{EFEFEF}38.51\%                     & \cellcolor[HTML]{EFEFEF}39.25\%          & \cellcolor[HTML]{EFEFEF}39.30\%          \\
                            & 5k                                                                             & \cellcolor[HTML]{EFEFEF}59.13\%                      & \cellcolor[HTML]{EFEFEF}61.10\%          & \cellcolor[HTML]{EFEFEF}62.25\%          & 61.30\%                                             & 62.87\%                                  & 63.87\%                                  & \cellcolor[HTML]{EFEFEF}60.80\%                     & \cellcolor[HTML]{EFEFEF}61.70\%          & \cellcolor[HTML]{EFEFEF}61.76\%          \\
                            & 10k                                                                            & \cellcolor[HTML]{EFEFEF}67.48\%                      & \cellcolor[HTML]{EFEFEF}69.70\%          & \cellcolor[HTML]{EFEFEF}70.98\%          & 69.24\%                                             & 71.02\%                                  & 71.91\%                                  & \cellcolor[HTML]{EFEFEF}68.04\%                     & \cellcolor[HTML]{EFEFEF}68.99\%          & \cellcolor[HTML]{EFEFEF}69.10\%          \\
                            & 20k                                                                            & \cellcolor[HTML]{EFEFEF}75.07\%                      & \cellcolor[HTML]{EFEFEF}77.46\%          & \cellcolor[HTML]{EFEFEF}79.01\%          & 76.81\%                                             & 78.40\%                                  & 79.94\%                                  & \cellcolor[HTML]{EFEFEF}74.41\%                     & \cellcolor[HTML]{EFEFEF}75.36\%          & \cellcolor[HTML]{EFEFEF}75.67\%          \\
\multirow{-5}{*}{SimpleNet} & 45k                                                                            & \cellcolor[HTML]{EFEFEF}\textbf{82.18\%}             & \cellcolor[HTML]{EFEFEF}\textbf{84.67\%} & \cellcolor[HTML]{EFEFEF}\textbf{86.38\%} & \textbf{82.72\%}                                    & \textbf{84.44\%}                         & \textbf{85.98\%}                         & \cellcolor[HTML]{EFEFEF}\textbf{80.48\%}            & \cellcolor[HTML]{EFEFEF}\textbf{81.46\%} & \cellcolor[HTML]{EFEFEF}\textbf{81.73\%} \\
                            & 1k                                                                             & 65.65\%                                              & 68.60\%                                  & 68.76\%                                  & \cellcolor[HTML]{EFEFEF}69.14\%                     & \cellcolor[HTML]{EFEFEF}71.32\%          & \cellcolor[HTML]{EFEFEF}71.49\%          & 72.19\%                                             & 73.20\%                                  & 73.41\%                                  \\
                            & 5k                                                                             & 76.33\%                                              & 79.60\%                                  & 79.41\%                                  & \cellcolor[HTML]{EFEFEF}79.79\%                     & \cellcolor[HTML]{EFEFEF}82.19\%          & \cellcolor[HTML]{EFEFEF}82.24\%          & 82.78\%                                             & 83.81\%                                  & 84.06\%                                  \\
                            & 10k                                                                            & 80.10\%                                              & 83.48\%                                  & 83.23\%                                  & \cellcolor[HTML]{EFEFEF}82.25\%                     & \cellcolor[HTML]{EFEFEF}84.83\%          & \cellcolor[HTML]{EFEFEF}84.69\%          & 85.09\%                                             & 86.13\%                                  & 86.38\%                                  \\
                            & 20k                                                                            & 81.91\%                                              & 85.21\%                                  & 85.40\%                                  & \cellcolor[HTML]{EFEFEF}84.84\%                     & \cellcolor[HTML]{EFEFEF}87.32\%          & \cellcolor[HTML]{EFEFEF}87.52\%          & 87.90\%                                             & 89.01\%                                  & 89.12\%                                  \\
\multirow{-5}{*}{ResNet-18} & 45k                                                                            & \textbf{84.28\%}                                     & \textbf{87.59\%}                         & \textbf{87.83\%}                         & \cellcolor[HTML]{EFEFEF}\textbf{86.88\%}            & \cellcolor[HTML]{EFEFEF}\textbf{89.43\%} & \cellcolor[HTML]{EFEFEF}\textbf{89.64\%} & \textbf{90.42\%}                                    & \textbf{91.44\%}                         & \textbf{91.76\%}                         \\
                            & 1k                                                                             & \cellcolor[HTML]{EFEFEF}63.60\%                      & \cellcolor[HTML]{EFEFEF}66.77\%          & \cellcolor[HTML]{EFEFEF}66.55\%          & 71.10\%                                             & 73.50\%                                  & 73.18\%                                  & \cellcolor[HTML]{EFEFEF}73.74\%                     & \cellcolor[HTML]{EFEFEF}74.73\%          & \cellcolor[HTML]{EFEFEF}74.99\%          \\
                            & 5k                                                                             & \cellcolor[HTML]{EFEFEF}74.59\%                      & \cellcolor[HTML]{EFEFEF}78.03\%          & \cellcolor[HTML]{EFEFEF}77.57\%          & 79.31\%                                             & 81.94\%                                  & 81.61\%                                  & \cellcolor[HTML]{EFEFEF}85.22\%                     & \cellcolor[HTML]{EFEFEF}86.11\%          & \cellcolor[HTML]{EFEFEF}86.60\%          \\
                            & 10k                                                                            & \cellcolor[HTML]{EFEFEF}79.63\%                      & \cellcolor[HTML]{EFEFEF}83.05\%          & \cellcolor[HTML]{EFEFEF}82.65\%          & 83.17\%                                             & 85.95\%                                  & 85.46\%                                  & \cellcolor[HTML]{EFEFEF}87.92\%                     & \cellcolor[HTML]{EFEFEF}88.84\%          & \cellcolor[HTML]{EFEFEF}89.29\%          \\
                            & 20k                                                                            & \cellcolor[HTML]{EFEFEF}83.46\%                      & \cellcolor[HTML]{EFEFEF}86.94\%          & \cellcolor[HTML]{EFEFEF}86.56\%          & 86.36\%                                             & 89.14\%                                  & 88.79\%                                  & \cellcolor[HTML]{EFEFEF}89.96\%                     & \cellcolor[HTML]{EFEFEF}90.97\%          & \cellcolor[HTML]{EFEFEF}91.30\%          \\
\multirow{-5}{*}{ResNet-34} & 45k                                                                            & \cellcolor[HTML]{EFEFEF}\textbf{85.17\%}             & \cellcolor[HTML]{EFEFEF}\textbf{88.73\%} & \cellcolor[HTML]{EFEFEF}\textbf{88.49\%} & \textbf{87.56\%}                                    & \textbf{90.15\%}                         & \textbf{90.26\%}                         & \cellcolor[HTML]{EFEFEF}\textbf{92.04\%}            & \cellcolor[HTML]{EFEFEF}\textbf{93.28\%} & \cellcolor[HTML]{EFEFEF}\textbf{93.25\%} \\ \bottomrule
\end{tabular}
\label{tab:attacks_substitute_arch_cinic}
\end{table*}

\begin{table*}[t]
\centering
\caption{Performance of substitute models with different architectures trained on the artificial data.}
\begin{tabular}{ccccccccccc}
\toprule
                            & \textbf{Target $\rightarrow$}                                                           & \multicolumn{3}{c}{\textbf{SimpleNet}}                                                                                                              & \multicolumn{3}{c}{\begin{tabular}[c]{@{}c@{}}\textbf{ResNet-34} \\ \textbf{(from scratch)}\end{tabular}}                                                  & \multicolumn{3}{c}{\begin{tabular}[c]{@{}c@{}}\textbf{ResNet-34} \\ \textbf{(transfer learning)}\end{tabular}}                                              \\ \cmidrule(lr){3-5} \cmidrule(lr){6-8} \cmidrule(lr){9-11}
\textbf{Substitute $\downarrow$}     &  \textit{QB}  & \textit{Joint Acc} & \textit{Acc}                                      & \textit{Fid}                                      & \textit{Joint Acc} & \textit{Acc}                                      & \textit{Fid}                                     & \textit{Joint Acc} & \textit{Acc}                                      & \textit{Fid}                                      \\ \midrule
                            & 1k                   & \cellcolor[HTML]{EFEFEF}21.27\%          & \cellcolor[HTML]{EFEFEF}22.65\%          & \cellcolor[HTML]{EFEFEF}23.10\%          & 23.34\%                                  & 24.39\%                                  & 24.58\%                                  & \cellcolor[HTML]{EFEFEF}22.93\%          & \cellcolor[HTML]{EFEFEF}23.48\%          & \cellcolor[HTML]{EFEFEF}23.57\%          \\
                            & 5k                   & \cellcolor[HTML]{EFEFEF}38.51\%          & \cellcolor[HTML]{EFEFEF}40.31\%          & \cellcolor[HTML]{EFEFEF}40.77\%          & 36.13\%                                  & 37.5\%                                   & 37.78\%                                  & \cellcolor[HTML]{EFEFEF}34.43\%          & \cellcolor[HTML]{EFEFEF}35.11\%          & \cellcolor[HTML]{EFEFEF}35.04\%          \\
                            & 10k                  & \cellcolor[HTML]{EFEFEF}45.87\%          & \cellcolor[HTML]{EFEFEF}47.75\%          & \cellcolor[HTML]{EFEFEF}48.61\%          & 41.56\%                                  & 43.02\%                                  & 43.49\%                                  & \cellcolor[HTML]{EFEFEF}37.66\%          & \cellcolor[HTML]{EFEFEF}38.28\%          & \cellcolor[HTML]{EFEFEF}38.28\%          \\
                            & 20k                  & \cellcolor[HTML]{EFEFEF}55.06\%          & \cellcolor[HTML]{EFEFEF}57.03\%          & \cellcolor[HTML]{EFEFEF}58.12\%          & 52.68\%                                  & 54.33\%                                  & 54.81\%                                  & \cellcolor[HTML]{EFEFEF}44.12\%          & \cellcolor[HTML]{EFEFEF}44.87\%          & \cellcolor[HTML]{EFEFEF}44.95\%          \\
\multirow{-5}{*}{SimpleNet} & 45k                  & \cellcolor[HTML]{EFEFEF}\textbf{64.88\%} & \cellcolor[HTML]{EFEFEF}\textbf{67.10\%} & \cellcolor[HTML]{EFEFEF}\textbf{68.35\%} & \textbf{65.83\%}                         & \textbf{67.57\%}                         & \textbf{68.34\%}                         & \cellcolor[HTML]{EFEFEF}\textbf{50.19\%} & \cellcolor[HTML]{EFEFEF}\textbf{50.99\%} & \cellcolor[HTML]{EFEFEF}\textbf{51.01\%} \\
                            & 1k                   & 63.28\%                                  & 66.20\%                                  & 65.86\%                                  & \cellcolor[HTML]{EFEFEF}64.13\%          & \cellcolor[HTML]{EFEFEF}66.06\%          & \cellcolor[HTML]{EFEFEF}66.32\%          & 66.05\%                                  & 66.83\%                                  & 67.10\%                                  \\
                            & 5k                   & 67.90\%                                  & 71.11\%                                  & 70.55\%                                  & \cellcolor[HTML]{EFEFEF}67.30\%          & \cellcolor[HTML]{EFEFEF}69.44\%          & \cellcolor[HTML]{EFEFEF}69.52\%          & 71.12\%                                  & 71.97\%                                  & 72.14\%                                  \\
                            & 10k                  & 69.93\%                                  & 73.04\%                                  & 72.86\%                                  & \cellcolor[HTML]{EFEFEF}71.38\%          & \cellcolor[HTML]{EFEFEF}73.72\%          & \cellcolor[HTML]{EFEFEF}73.40\%          & 69.58\%                                  & 70.46\%                                  & 70.63\%                                  \\
                            & 20k                  & 73.43\%                                  & 76.72\%                                  & 76.42\%                                  & \cellcolor[HTML]{EFEFEF}73.96\%          & \cellcolor[HTML]{EFEFEF}76.31\%          & \cellcolor[HTML]{EFEFEF}76.19\%          & 68.08\%                                  & 69.01\%                                  & 69.16\%                                  \\
\multirow{-5}{*}{ResNet-18} & 45k                  & \textbf{75.69\%}                         & \textbf{78.88\%}                         & \textbf{78.78\%}                         & \cellcolor[HTML]{EFEFEF}\textbf{74.94\%} & \cellcolor[HTML]{EFEFEF}\textbf{77.10\%} & \cellcolor[HTML]{EFEFEF}\textbf{77.28\%} & \textbf{72.96\%}                         & \textbf{73.81\%}                         & \textbf{74.00\%}                         \\
                            & 1k                   & \cellcolor[HTML]{EFEFEF}62.11\%          & \cellcolor[HTML]{EFEFEF}65.16\%          & \cellcolor[HTML]{EFEFEF}64.37\%          & 63.38\%                                  & 65.69\%                                  & 65.44\%                                  & \cellcolor[HTML]{EFEFEF}65.39\%          & \cellcolor[HTML]{EFEFEF}66.20\%          & \cellcolor[HTML]{EFEFEF}66.48\%          \\
                            & 5k                   & \cellcolor[HTML]{EFEFEF}66.46\%          & \cellcolor[HTML]{EFEFEF}69.58\%          & \cellcolor[HTML]{EFEFEF}69.02\%          & 67.03\%                                  & 69.33\%                                  & 69.16\%                                  & \cellcolor[HTML]{EFEFEF}66.09\%          & \cellcolor[HTML]{EFEFEF}66.86\%          & \cellcolor[HTML]{EFEFEF}67.09\%          \\
                            & 10k                  & \cellcolor[HTML]{EFEFEF}71.83\%          & \cellcolor[HTML]{EFEFEF}75.15\%          & \cellcolor[HTML]{EFEFEF}74.46\%          & 72.62\%                                  & 74.89\%                                  & 74.86\%                                  & \cellcolor[HTML]{EFEFEF}68.83\%          & \cellcolor[HTML]{EFEFEF}69.73\%          & \cellcolor[HTML]{EFEFEF}69.82\%          \\
                            & 20k                  & \cellcolor[HTML]{EFEFEF}75.39\%          & \cellcolor[HTML]{EFEFEF}78.75\%          & \cellcolor[HTML]{EFEFEF}78.17\%          & 73.59\%                                  & 76.05\%                                  & 75.81\%                                  & \cellcolor[HTML]{EFEFEF}70.75\%          & \cellcolor[HTML]{EFEFEF}71.65\%          & \cellcolor[HTML]{EFEFEF}71.84\%          \\
\multirow{-5}{*}{ResNet-34} & 45k                  & \cellcolor[HTML]{EFEFEF}\textbf{77.83\%} & \cellcolor[HTML]{EFEFEF}\textbf{81.03\%} & \cellcolor[HTML]{EFEFEF}\textbf{80.88\%} & \textbf{74.94\%}                         & \textbf{77.17\%}                         & \textbf{77.44\%}                         & \cellcolor[HTML]{EFEFEF}\textbf{75.12\%} & \cellcolor[HTML]{EFEFEF}\textbf{75.97\%} & \cellcolor[HTML]{EFEFEF}\textbf{76.38\%} \\ \bottomrule
\end{tabular}
\label{tab:attacks_substitute_arch_artificial}
\end{table*}

\subsection{Artificial Dataset Generation}
\label{appendix:artifical_dataset_generation}
The artificial dataset for the data-free attack was generated using a pre-trained stable diffusion model\footnote{\url{https://huggingface.co/stabilityai/stable-diffusion-2-1}}. For each class of the original dataset (CIFAR-10), we create text prompts to generate images. Each prompt consists of two parts: positive and negative. The positive prompt corresponds to inclusion criteria, and the negative prompt corresponds to exclusion criteria. Positive prompts usually include either a class label ("bird") or its subcategory ("swan"). Using subcategories makes the artificial data more diverse and helps to approximate the original data better. However, if no information is provided on the subcategories appearing in the original dataset, using them can be even misleading for a substitute model. Negative prompts aim to correct mistakes a diffusion model may make. For instance, they can prevent occurrences of bad anatomy or visual artefacts. 

\begin{table*}[t]
\centering
\footnotesize
\caption{Prompts used to generate the artificial dataset.}
\resizebox{\textwidth}{!}{\begin{tabular}{p{1.6cm}p{8.5cm}p{7cm}}
\toprule
\textbf{Class}      &\textbf{ Positive prompt}                                                                                                                                                                                                                                                                                                                                                                                                                                                                                                                                                                                                            & \textbf{Negative prompt}                                                                                                                                                                                                                                                                                                                                                                         \\ \midrule
airplane   & plane photo                                                                                                                                                                                                                                                                                                                                                                                                                                                                                                                                                                                                                                       & 3d, grid, deformed, ugly, mutation, mutated, blurry background, bokeh, multiple images, illustration, cropped, partial view, jpeg artifacts, grayscale                                                                                                                                                                                                                                  \\ \addlinespace
automobile & car photo | automobile photo                                                                                                                                                                                                                                                                                                                                                                                                                                                                                                                                                                                                                & 3d, grid, deformed, ugly, mutation, mutated, blurry background, bokeh, multiple images, illustration, cropped, partial view, jpeg artifacts, grayscale                                                                                                                                                                                                                                  \\ \addlinespace
bird       & cassowary photo | ostrich photo | emu photo |  kiwi bird photo | owl photo | hawk photo | grebe photo | loon photo | duck photo | pheasant photo | tern photo | hummingbird photo | hen photo | rooster photo | swan photo | goose photo | parrot photo | bustard photo | tit photo | sparrow photo | woodpecker photo | pigeon photo | cuckoo photo | raven photo | oriole photo | warbler photo | chickadee photo | starling photo | dove photo | finch photo | nuthatch photo | bird photo & 3d, bad anatomy, duplicated eyes, no eyes, extra eyes, grid, extra limbs, close up, deformed, ugly, mutation, mutated, blurry background, bokeh, multiple birds, multiple images, illustration, cropped, partial view, duplicated limbs,  jpeg artifacts, missing limb, floating limbs, disconnected limbs, black and white, two heads                                                  \\ \addlinespace
cat        & cat photo                                                                                                                                                                                                                                                                                                                                                                                                                                                                                                                                                                                                                                         & 3d, bad anatomy, duplicated eyes, no eyes, extra eyes, grid, extra limbs, close up, deformed, ugly, mutation, mutated, blurry background, bokeh, multiple cats, multiple images, illustration, cropped, partial view, duplicated limbs,  jpeg artifacts, missing limb, floating limbs, disconnected limbs, black and white                                                              \\ \addlinespace
deer       & deer photo                                                                                                                                                                                                                                                                                                                                                                                                                                                                                                                                                                                                                                        & 3d, bad anatomy, duplicated head, missing head, extra head, grid, extra limbs, close up, deformed, ugly, mutation, mutated, blurry background, bokeh, multiple deers, multiple images, illustration, cropped, partial view, duplicated limbs,  jpeg artifacts, missing limb, floating limbs, disconnected limbs, black and white, grayscale, painting, watermark, signature, two heads  \\ \addlinespace
dog        & dog photo                                                                                                                                                                                                                                                                                                                                                                                                                                                                                                                                                                                                                                        & 3d, bad anatomy, duplicated eyes, no eyes, extra eyes, grid, extra limbs, close up, deformed, ugly, mutation, mutated, blurry background, bokeh, multiple dogs, multiple images, illustration, cropped, partial view, duplicated limbs,  jpeg artifacts, missing limb, floating limbs, disconnected limbs, black and white                                                              \\ \addlinespace
frog       & brown frog photo | green frog photo                                                                                                                                                                                                                                                                                                                                                                                                                                                                                                                                                                                                       & 3d, bad anatomy, duplicated eyes, no eyes, extra eyes, grid, extra limbs, close up, deformed, ugly, mutation, mutated, blurry background, bokeh, multiple frogs, multiple images, illustration, cropped, partial view, duplicated limbs,  jpeg artifacts, missing limb, floating limbs, disconnected limbs                                                                              \\ \addlinespace
horse      & black horse photo | gray horse photo | chestnut horse photo | bay horse photo | dun horse photo                                                                                                                                                                                                                                                                                                                                                                                                                                                                                                                          & 3d, bad anatomy, duplicated head, missing head, extra head, grid, extra limbs, close up, deformed, ugly, mutation, mutated, blurry background, bokeh, multiple horses, multiple images, illustration, cropped, partial view, duplicated limbs,  jpeg artifacts, missing limb, floating limbs, disconnected limbs, black and white, grayscale, painting, watermark, signature, two heads \\ \addlinespace
ship       & watercraft photo | ship photo | sailboat photo                                                                                                                                                                                                                                                                                                                                                                                                                                                                                                                                                                                       & 3d, grid, deformed, ugly, mutation, mutated, blurry background, bokeh, multiple images, illustration, cropped, partial view, jpeg artifacts, grayscale                                                                                                                                                                                                                                  \\ \addlinespace
truck      & truck photo                                                                                                                                                                                                                                                                                                                                                                                                                                                                                                                                                                                                                                       & 3d, grid, deformed, ugly, mutation, mutated, blurry background, bokeh, multiple images, illustration, cropped, partial view,  jpeg artifacts, grayscale                                                                                                                                                                                                                                 \\ \bottomrule
\end{tabular}
}
\label{tab:prompts}
\end{table*}

\Cref{tab:prompts} shows positive and negative prompts used to generate images of each class. Sometimes, the model was biased towards generating very similar images for the same positive prompt. For instance, the prompt "bird photo" rendered birds of the same size and colour. Hence, we used a list of different bird families to generate representatives for them, making the dataset more diverse. We also replaced "airplane" with "plane" in the positive prompt because the model always generated the same type of aircraft, up in the air, for the "airplane" prompt. In contrast, "plane" images were more diverse and contained both flying and still vehicles. We observed similar behaviour for the "ship" class and addressed it by adding two additional prompts. "Frog" and "horse" classes lacked diversity in colours, so we asked the model explicitly to make them more diverse. 

\end{document}